\documentclass[a4paper,fleqn]{cas-dc}
\usepackage{amsmath,amsfonts}
\usepackage{algorithmic}
\usepackage{algorithm}
\usepackage{array}
\usepackage[caption=false,font=normalsize,labelfont=sf,textfont=sf]{subfig}
\usepackage{textcomp}
\usepackage{stfloats}
\usepackage{url}
\usepackage{verbatim}
\usepackage{graphicx}
\usepackage{cite}
\usepackage{times}
\usepackage{soul}
\usepackage{url}
\usepackage{environ}
\usepackage{hyperref}
\usepackage[utf8]{inputenc}
\usepackage{graphicx}
\usepackage{amsmath}
\usepackage{booktabs}
\usepackage{algorithm}
\usepackage{array}
\usepackage{adjustbox}
\usepackage{makecell}
\usepackage{tabularx}
\usepackage{empheq}

\usepackage[flushleft]{threeparttable}

\newcolumntype{P}[1]{>{\centering\arraybackslash}p{#1}}
\newcolumntype{C}[1]{>{\centering\arraybackslash}m{#1}}

\usepackage{enumitem,picinpar}
\usepackage[framemethod=tikz]{mdframed}
\usepackage{amsmath}
\usepackage{mathtools,nccmath}
\usepackage{threeparttable}
\usepackage{tablefootnote}
\usepackage{longtable}

\newlength\widest
\makeatletter
\NewEnviron{ldescription}{%
  \vbox{%
    \global\setlength\widest{0pt}%
    \def\item[##1]{%
      \settowidth\@tempdima{\textbf{##1}}%
      \ifdim\@tempdima>\widest\global\setlength\widest{\@tempdima}\fi%
    }%
    \setbox0=\hbox{\BODY}%
  }
  \begin{description}[
    leftmargin=\dimexpr\widest+0.5em\relax,
    labelindent=0pt,
    labelwidth=\widest]
  \BODY
  \end{description}%
}

\usepackage[square,comma,numbers]{natbib}

\def\tsc#1{\csdef{#1}{\textsc{\lowercase{#1}}\xspace}}
\tsc{WGM}
\tsc{QE}

\emergencystretch=\maxdimen
\begin{document}
\let\WriteBookmarks\relax
\def\floatpagepagefraction{1}
\def\textpagefraction{.001}

\shorttitle{A Recent Survey of Heterogeneous Transfer Learning}    

\shortauthors{R. Bao et al.}  

\title [mode = title]{A Recent Survey of Heterogeneous Transfer Learning}  



%

\author[1]{Runxue Bao}[]
\fnmark[*]



\ead{runxue.bao@gehealthcare.com}


\credit{Conceptualization, Methodology, Investigation, Validation, Writing - original draft}

\affiliation[1]{organization={GE Healthcare},
            city={Bellevue},
            state={WA},
            postcode={98004},
            country={USA}}

\author[2]{Yiming Sun}[]

\fnmark[*]
\credit{Conceptualization, Methodology, Investigation, Formal analysis, Writing - original draft}

\ead{yis108@pitt.edu}



\affiliation[2]{organization={University of Pittsburgh},
            city={Pittsburgh},
            state={PA},
            postcode={15260}, 
            country={USA}}
            
\affiliation[3]{organization={Microsoft Research Asia},
            state={Beijing},
            postcode={100080},
            country={China}}

\affiliation[4]{organization={Hong Kong University of Science and Technology},
            city={Hong Kong},
            postcode={999077}, 
            country={China}}

\author[2]{Yuhe Gao}[]
\ead{yug51@pitt.edu}
\credit{Visualization, Writing - review and editing}

\author[3]{Jindong Wang}[]
\ead{jindong.wang@microsoft.com}
\credit{Validation, Writing - review and editing}

\author[4]{Qiang Yang}[]
\ead{qyang@cse.ust.hk}
\credit{Validation, Writing - review and editing}

\author[2]{Zhi-Hong Mao}[]
\ead{zhm4@pitt.edu}
\credit{Writing - review and editing}

\author[2]{Ye Ye}[orcid=0000-0002-1138-9846]
\fnmark[**]
\ead{yey5@pitt.edu}
\credit{Conceptualization, Funding acquisition, Supervision, Validation, Writing - review and editing}

\cortext[1]{Equal Contribution}
\cortext[2]{Corresponding author}



\begin{abstract}
The application of transfer learning, leveraging knowledge from source domains to enhance model performance in a target domain, has significantly grown, supporting diverse real-world applications. Its success often relies on shared knowledge between domains, typically required in these methodologies. Commonly, methods assume identical feature and label spaces in both domains, known as homogeneous transfer learning. However, this is often impractical as source and target domains usually differ in these spaces, making precise data matching challenging and costly. Consequently, heterogeneous transfer learning (HTL), which addresses these disparities, has become a vital strategy in various tasks.
In this paper, we offer an extensive review of over 60 HTL methods, covering both data-based and model-based approaches. We describe the key assumptions and algorithms of these methods and systematically categorize them into instance-based, feature representation-based, parameter regularization, and parameter tuning techniques. Additionally, we explore applications in natural language processing, computer vision, multimodal learning, and biomedicine, aiming to deepen understanding and stimulate further research in these areas. Our paper includes recent advancements in HTL, such as the introduction of transformer-based models and multimodal learning techniques, ensuring the review captures the latest developments in the field. We identify key limitations in current HTL studies and offer systematic guidance for future research, highlighting areas needing further exploration and suggesting potential directions for advancing the field.
\end{abstract}


\maketitle

\section{Introduction}







In recent decades, the field of machine learning has experienced remarkable achievements across diverse domains of application. Notably, the substantial progress made in machine learning can be attributed to the extensive utilization of abundant labeled datasets in the era of big data. Nonetheless, the acquisition of labeled data can present challenges in terms of cost or feasibility within certain practical scenarios. To address this issue, transfer learning \cite{yang_zhang_dai_pan_2020, pan2010survey, zhuang2020comprehensive, niu2020decade, weiss2016survey} has emerged as a promising technique for enhancing model performance in a target domain by leveraging knowledge transfer from one or more source domains. The source domain typically offers a more accessible or economical means of obtaining labeled data. This notion exhibits conceptual similarities to the transfer learning paradigm observed in psychological literature, where the aim is to generalize experiences from prior activities to new ones. For instance, the knowledge (e.g., pitch relationships, harmonic progressions, and musical structures) acquired from playing violins can be applied to the task of playing pianos, serving as a practical illustration of transfer learning. The effectiveness of transfer learning crucially hinges on the relevance between the new task and past tasks.

Typically, transfer learning is divided into two main categories: homogeneous transfer learning and heterogeneous transfer learning (HTL). The former pertains to scenarios where the source and target domains have matching feature and label spaces. 
However, real-world applications frequently involve disparate feature spaces and, occasionally, dissimilar label spaces between the source and target domains. Unfortunately, in these scenarios, collecting source domain data that seamlessly aligns with the target domain's feature space can prove infeasible or prohibitively expensive. Moreover, as new data and domains emerge, HTL facilitates models to continuously adapt and remain up-to-date without beginning from scratch. Consequently, researchers have directed significant attention towards investigating HTL techniques, which have shown promise across various tasks \cite{CDLS, SHDA-RF, HeMap, HDAMA}.

Previous literature reviews have predominantly focused on homogeneous transfer learning approaches. 
Several surveys \cite{zhuang2020comprehensive,niu2020decade, weiss2016survey,zhang2022transfer,agarwal2021survey} have systematically categorized and assessed a wide spectrum of transfer learning techniques, taking into account various aspects such as algorithmic categories and application scenarios. 
An emerging trend is conducting literature reviews on technologies that combine transfer learning with other machine learning techniques, such as deep learning \cite{tan2018survey,iman2023review}, reinforcement learning \cite{liang2019survey, taylor2009transfer, zhu2023reinforcement}, and federated learning \cite{saha2021federated, hallaji2022federated}.
Beyond algorithm-centric surveys, certain reviews have concentrated specifically on applications in computer vision (CV) \cite{shao2014transfer, patel2015visual, wang2018deep,cook2013transfer}, natural language processing (NLP) \cite{alyafeai2020survey, liu2019survey, ruder2019transfer}, medical image analysis \cite{yu2022transfer,sufian2020survey}, and wireless communication \cite{nguyen2021transfer, wong2022transfer}.


While there exist three surveys \cite{day2017survey, friedjungova2017asymmetric,HTL2024} on HTL, the first two surveys primarily cover approaches proposed before 2017.
The third survey \cite{HTL2024} is a recent one, but focused only on features-based algorithms, a subset of the HTL methods. 
All of them fail to incorporate the latest advancements in this area, especially the advert of transformer \cite{vaswani2017attention} and its descendants, such as 
\textbf{B}idirectional \textbf{E}ncoder \textbf{R}epresentations from \textbf{T}ransformers (BERT) \cite{BERT} and \textbf{G}enerative \textbf{P}re-trained \textbf{T}ransformer (GPT) \cite{GPT-1}. Since 2017, the field of HTL has continued to flourish with ongoing research. Specifically, large-scale foundation models are publicly available, exhibiting significant potential to provide a robust and task-agnostic starting point for transfer learning applications. Leveraging HTL not only enhances model performance on target tasks by initiating with pre-existing knowledge but also significantly reduces training time and resource usage through fine-tuning of pre-trained models. Furthermore, another notable advancement is the embrace of multi-modality, where knowledge from different domains is combined to enhance learning outcomes \cite{liang2022foundations, zhen2020deep}. Multimodal learning has shown tremendous promise in handling data from diverse modalities like images, text, and audio, which is pivotal in tasks such as image captioning, visual question answering, and cross-modal retrieval.  In summary, HTL is of paramount importance as it substantially enhances the performance, adaptability, and efficiency of machine learning models across an extensive range of applications. Since there has been a notable absence of subsequent summarization efforts to capture the advancements in this area, to fill the gap, we present an exhaustive review of the state-of-the-art in HTL, with a focus on recent breakthroughs.

\hfill
\\
\noindent \textbf{Contributions.} This survey significantly contributes to the field of HTL by providing an extensive overview of methodologies and applications\footnote{The papers reviewed in the survey, along with associated resources including code and datasets, can be accessed at \url{https://github.com/ymsun99/Heterogeneous-Transfer-Learning}.}, and offering detailed insights to guide future research.  The key contributions are:
\begin{enumerate}[leftmargin=0.2in]

\item This paper provides an extensive review of more than 60 HTL methods, detailing their underlying assumptions, and key algorithms. It systematically categorizes these methods into data-based and model-based approaches, offering insights into different HTL strategies, including instance-based, feature representation-based, parameter regularization, and parameter tuning.
\item The survey includes recent advancements in HTL, such as the introduction of transformer-based models and multimodal learning techniques, ensuring the review captures the latest developments in the field.
\item The survey identifies key limitations in current HTL studies and offers systematic guidance for future research. It highlights areas needing further exploration and suggests potential directions for advancing the field.

\end{enumerate}

\noindent \textbf{Organization.}
We organize the rest of the paper as follows. Firstly, we introduce notations and problem definitions in Section \ref{prelim}. Secondly, we provide an overview of data-based HTL methods in Section \ref{data-based}, including instance-based and feature representation-based approaches. Thirdly, we discuss model-based methods in Section \ref{model-based}. Lastly, we delve into methods in application scenarios in Section \ref{application}. Finally, we present the concluding remarks of the paper.


\begin{figure*}[htbp]
\centerline{\includegraphics[width=170mm]{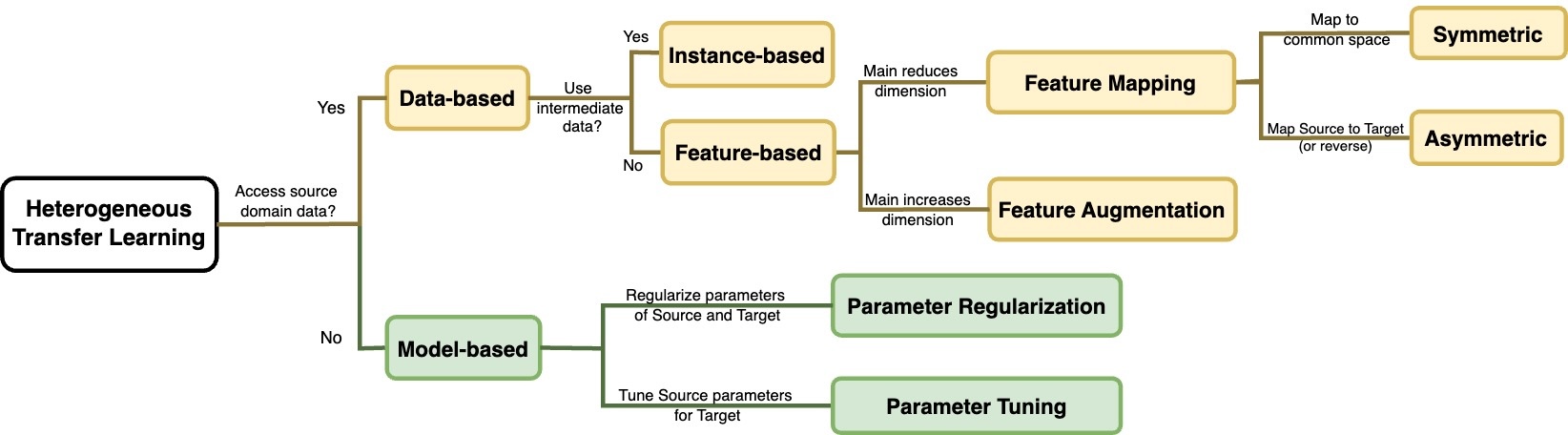}}
\caption{The summary of approaches in heterogeneous transfer learning.}
\label{fig1_sub}
\end{figure*}

\section{Preliminary}
\label{prelim}

\subsection{Notations and Problem Definitions}

\noindent \textbf{Notations.} To simplify understanding, we provide a summary of notations in the following list.
\begin{ldescription}
\item[$D_S$] Source Domain.
\item[$D_T$] Target Domain.
\item[$d_S$] Feature size of the source domain.
\item[$d_T$] Feature size of the target domain.
\item[$n_S$] Instance size of the source domain.
\item[$n_T$] Instance size of the target domain.
\item[$\mathcal{X}_S$] Feature space of the source domain.
\item[$x_S \in \mathbb{R}^{d_S}$] Feature vector of one instance in the source domain.
\item[$X_S \in \mathbb{R}^{n_S \times d_S}$] Data matrix of all instances in the source domain.
\item [$\mathcal{Y}_{S}$] Label space of the source domain.
\item[$y_S \in \mathbb{R}^{n_S}$] Labels of all instances in the source domain.
\item [$\mathcal{X}_T$] Feature space of the target domain.
\item[$x_T \in \mathbb{R}^{d_T}$] Feature vector of one instance in the target domain.
\item[$X_T \in \mathbb{R}^{n_T \times d_T}$] Data matrix of all instances in the target domain.
\item[$y_T \in \mathbb{R}^{n_T}$] Labels of all instances in the target domain.
\item[$\mathcal{Y}_{T}$] Label space of the target domain.
\item[$R(\cdot)$] Regularization function.
\item[$\mathcal{L}(\cdot)$] Objective function.
\end{ldescription}

\noindent \textbf{Problem Definitions.}
In this survey, a domain $D$ comprises a feature space $\mathcal{X}$ and a marginal probability distribution $P\left(x\right)$ where $x \in \mathcal{X}$. For a given specific domain $D=\{\mathcal{X}, P\left(x\right)\}$, a task $\mathcal{T}$ consists a label space $\mathcal{Y}$ and an objective predictive function $P\left(y\mid x\right)$. Source domain data is denoted as $D_S=\{X_S, y_S\}=\{\left(x_{S,1}, y_{S,1}\right), \dots, \left(x_{S,n_S}, y_{S,n_S}\right)\}$ where $x_{S,i} \in \mathcal{X_S}$ and $y_{S,i} \in \mathcal{Y_S}$, and similarly, target domain data is denoted as \\
$D_T=\{X_T, y_T\}=\{\left(x_{T,1}, y_{T,1}\right), \dots, \left(x_{T,n_T}, y_{T,n_T}\right)\}$ where $x_{T,i} \in \mathcal{X}_T$ and $y_{T,i} \in \mathcal{Y}_T$. In most cases, $0\leq n_T \ll n_S$.

Given source domain data $D_S$ and task $\mathcal{T}_S$, and target domain $D_T$ and task $\mathcal{T}_T$,
transfer learning, in this context, involves leveraging the knowledge from $D_S$ and $\mathcal{T}_S$ to enhance the learning of the objective predictive function $P_T\left(y\mid x\right)$ in $D_T$, where $D_S \neq D_T$ or $\mathcal{T}_S \neq \mathcal{T}_T$. Specifically, the condition $D_S \neq D_T$ indicates differences in either the feature spaces, $\mathcal{X}_S \neq \mathcal{X}_T$, or marginal distributions, $P_S\left(x\right) \neq P_T\left(x\right)$. Similarly, the condition $\mathcal{T}_S \neq \mathcal{T}_T$ implies disparities in either the label spaces $\mathcal{Y}_S \neq \mathcal{Y}_T$ or the objective functions $P_S\left(y\mid x\right) \neq P_T\left(y\mid x\right)$. These differences distinguish between homogeneous and heterogeneous transfer learning. In homogeneous transfer learning, feature spaces $\mathcal{X}$ and label spaces $\mathcal{Y}$ are identical, while marginal distributions $P(x)$ and objective functions $P(y|x)$ can differ. Conversely, heterogeneous transfer learning, which is the primary focus of this survey, pertains to scenarios where either $\mathcal{X}_S \neq \mathcal{X}_T$ or $\mathcal{Y}_S \neq \mathcal{Y}_T$.

Furthermore, within the realm of transfer learning, domain adaptation \cite{wang2018deep, farahani2021brief, csurka2017domain, wilson2020survey} is a subset characterized by $\mathcal{T}_S = \mathcal{T}_T$ and $D_S \neq D_T$. However, it is important to note that the terms ``domain adaptation'' and ``transfer learning'' are often used interchangeably in the literature.

\subsection{Learning Scenarios}


In HTL, the choice of methods is heavily influenced by the availability of labeled data in source and target domains. This section delves into three primary scenarios, each defined by the presence or absence of labeled data: (1) both source and target domains possess labeled data, though the target domain is likely to exhibit significant label scarcity; (2) only source domain has labels; and (3) an entirely unsupervised setting, where both domains do not have labels. 
These scenarios each bring forth distinct challenges and objectives, demanding specialized approaches to efficiently harnessing available information and enabling knowledge transfer.

\paragraph*{Source Labeled, Target Labeled:}
In this scenario, both the source and target domains possess labeled data. However, the target domain often lacks sufficient labeled data, which is a significant challenge. To address this, the methods in this category often use semi-supervised settings \cite{WOS:000899419900068} for the target domain. These settings comprise a limited amount of labeled data complemented by a substantial volume of unlabeled target data. 
The goal is to use the labeled data from both domains, along with the unlabeled target data, to improve learning in the target domain.

\paragraph*{Source Labeled, Target Unlabeled:}
In this specific scenario, labeled information is available exclusively from the source domain, leaving the target domain without labeled data. The challenge here involves utilizing the labeled source data effectively to make accurate predictions for the instances in the target domain.

\paragraph*{Unsupervised Transfer Learning:}
Unsupervised transfer learning addresses scenarios where instances in both the source and target domains are unlabeled. The primary objective in this context is to harness meaningful and transferable knowledge from a source domain to enhance learning in a target domain, notwithstanding the lack of labeled data.

\subsection{Data-based vs. Model-based }

The methodologies outlined in our survey can be broadly divided into two major categories: data-based methods, as covered in Section \ref{data-based}, and model-based methods, elaborated upon in Section \ref{model-based}. Figure \ref{fig1_sub} and Table \ref{methods} provides an overview.

Data-based methods involve the transfer of \textit{either the original data or their transformed features} to a target domain, allowing the target model to be trained with this augmented data, thereby enriching the available data within the target domain. Conversely, model-based methods center around constructing models and learning their parameters exclusively within the source domain. By adapting \textit{both the model structure and parameters of a source model}, the target models inherit the underlying insights from the prior knowledge in the source domain, consequently leading to enhanced performances.

Delving deeper, the data-based section distinguishes between instance-based methods in Section \ref{instance-based} and feature representation-based ones in Section \ref{feature-based}. Instance-based methods utilize 
\textit{intermediate data} 
that relates to both source and target domains, effectively serving as a bridge between them. In contrast, feature representation-based methods employ techniques such as feature mapping or feature augmentation to align the features of both domains, transforming them into a shared space without involving additional data.

In the model-based part, methods are also further classified into parameter-regularization in Section \ref{regularization} and parameter-tuning methods in Section \ref{tuning}. In the former category, the objective function integrates regularization techniques to control parameter differences between both models. Target models in this category begin with random initialization and are trained on target tasks. During training, they are constrained to ensure that their parameters do not significantly diverge from those of the source models.


Conversely, the latter category involves initializing target models using parameters from source models and subsequently refining them through fine-tuning on specific target tasks.



\begin{table*}[!t]

\caption{The summary of important references for different types of methods.}
\center
\begin{tabular}{cccc}
\toprule
\multicolumn{3}{c}
{\textbf{Method}}                                                                                                                     & \textbf{Important References  }       \\ \midrule
\multicolumn{1}{c}{\multirow{3}{*}{Data-based}}  & \multicolumn{2}{c}{Instance-based}                                                         &\cite{TTL, HTLIC, DHTL, OHKT, OHTHE, CDLS}                   \\ \cline{2-4} 
\multicolumn{1}{c}{}                             & \multicolumn{1}{c}{\multirow{2}{*}{Feature-based}} & Feature mapping                       &             \cite{ITML, brain2011what, SHDA-RF, SHFR, HeMap, HDAMA, DACoM, LPJT, ICDM, CDSPP, STN, SCT, HDAPA, HANDA, FSR, SSKMDA, TNT}      \\ \cline{3-4} 
\multicolumn{1}{c}{}                             & \multicolumn{1}{c}{}                               & Feature augmentation & \cite{HFA, SHFA, DCA, KPDA, SymGAN, HTLA, NMF, DeepMCA}  \\ \hline
\multicolumn{1}{c}{\multirow{2}{*}{Model-based}} & \multicolumn{2}{c}{Parameter Regularization}                                                            &  \cite{DTNs, REFORM}                 \\ \cline{2-4} 
\multicolumn{1}{c}{}                             & \multicolumn{2}{c}{Parameter Tuning}                                                       &  \cite{BERT, GPT-1, GPT-2, GPT-3, GPT-4, shen2021partial, lee2022surgical, BioBERT, wang2020generalizing}                 \\ 
\bottomrule
\end{tabular}             \label{methods}             
\end{table*}

\section{Data-based Method}
\label{data-based}

In transfer learning, data-based methods seek to integrate additional data instances that are not solely restricted to the target domain. These methods encompass instances from source domains and, where applicable, intermediate domains, as is especially pertinent in instance-based approaches. In HTL, the core strategy of these methods involves aligning the feature spaces that originate from both the source and target domains. This alignment fosters the creation of a unified, common space conducive to the integration of augmented information from all respective domains. By doing so, data-based methods significantly enrich the learning process, offering a substantial potential to boost models' adaptability and performance in varied scenarios.

\subsection{Instance-based Method}
\label{instance-based}

\begin{figure}[htbp]
\centerline{\includegraphics[width=90mm]{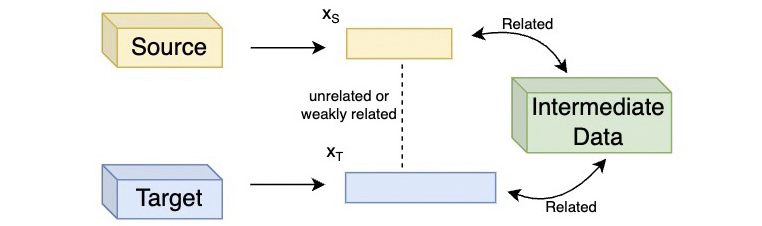}}
\caption{Instance-based method.}
\label{fig2}
\end{figure}

To establish a connection between heterogeneous source and target domains, it is intuitive to incorporate additional information to explore the latent relationships between these two feature spaces $X_S$ and $X_T$. Methods capitalizing on such supplementary information are classified under instance-based approaches. The supplementary information is termed as intermediate data $X_I$. Intermediate data, as shown in Figure \ref{fig2}, act as a bridge between the unrelated or weakly related source and target domains. The intermediate data shares relevance or characteristics with both source and target domains, thereby facilitating the discovery of underlying patterns and relationships between them. 


Instance-based methods draw inspiration from Multi-View Learning, where data instances are represented by multiple distinct feature representations or ``views''. Each view captures different facets or perspectives of the data, thereby providing a multifaceted understanding of the instances. In the context of intermediate data, one view may share homogeneous features with the source domain data, while another view shares the same characteristics with the target domain data. For example, in scenarios involving disparate data types like text and images, images with text annotations can serve as the intermediate data. With instance-based methods, the essence of knowledge transfer lies in the propagation of information from the source domain $X_S$, channeled through the intermediate domain $X_I$, ultimately reaching and enriching the target domain $X_T$ as shown in,
\begin{equation}
\label{eq:instance}
    X_S \longrightarrow X_I \longrightarrow X_T ~.
\end{equation}
We delve deeper into the exploration of intermediate data utilization through the following illustrative examples. 

\paragraph*{TTL:} \textbf{T}ransitive \textbf{T}ransfer \textbf{L}earning (TTL) \cite{TTL} introduces intermediate domain data $X_I$. This intermediate data is strategically designed to share distinct common factors with both the source domain $X_S$ and target domain $X_T$. TTL employs non-negative matrix tri-factorization (NMTF) on $X_S$, $X_I$ and $X_T$, which is formulated as $\left\| X - F A G^{\top} \right\|$. This approach is applied concurrently across the three domains. In this formulation, $X \in \mathbb{R}^{d\times n}$ represents the data matrix. Given The variables $p$ and $c$ represent the number of feature clusters and instance clusters, $F \in \mathbb{R}^{d\times p}$, $G \in \mathbb{R}^{n\times c}$, and $A\in\mathbb{R}^{p\times c}$ correspond to feature clusters, instance clusters, and the associations between feature clusters and instance clusters respectively. TTL's core mechanism involves feature clustering through NMTF, resulting in two interrelated feature representations. Knowledge transfer occurs by propagating label information from the source domain to the target domain. This process uses two pairs of coupled feature representations: one links the source and intermediate domains, and the other connects the target and intermediate domains.

\paragraph*{HTLIC:} In some cases, directly obtaining corresponding pairs between target and source domains can be challenging. Instead of relying on such pairs, the \textbf{H}eterogeneous \textbf{T}ransfer \textbf{L}earning for \textbf{I}mage \textbf{C}lassification (HTLIC) method \cite{HTLIC} enriches the representation of target images with semantic concepts extracted from auxiliary source documents. HTLIC incorporates intermediate data, which are auxiliary images that have been annotated with text tags sourced from the social Web, effectively establishing a bridge between image (the target domain) and text (the source domain).
HTLIC employs two matrices, specifically denoted as $G$ and $F$, which capture correlations between images and tags, as well as text and tags, respectively. Unlike traditional class labels, these tags encapsulate semantic representations that describe specific attributes or characteristics of data instances. Through the application of matrix bi-factorization techniques and the minimization of the objective function,
\begin{equation}
    \min _{U, V, W} \lambda\left\|G-U V^{\top}\right\|_F^2+(1-\lambda)\left\|F-W V^{\top}\right\|_F^2+R(U, V, W) ~,
\end{equation} 
where $U$, $V$, and $W$ represent the latent representations for target image instances, intermediate tags, and source document instances respectively, HTLIC learns the latent representation $U$. Following that, HTLIC incorporates the obtained latent representations $U$ into the target instances, resulting in the generation of transformed features $\hat{X}_T = X_TU$. 

\paragraph*{DHTL:} Inspired by the success of deep neural networks (DNNs) in transfer learning, in \cite{DHTL}, a \textbf{D}eep semantic mapping model for \textbf{H}eterogeneous multimedia \textbf{T}ransfer \textbf{L}earning (DHTL) method utilizes a specialized form of intermediate data called co-occurrence data. This method utilizes a specialized form of intermediate data known as co-occurrence data, which consists of instance pairs—one from the source domain and one from the target domain, such as text-to-image pairs and multilingual text pairs. DHTL is proposed to integrate auto-encoders with multiple layers to jointly learn the domain-specific networks and the shared inter-domain representation using co-occurrence data. 
To facilitate the alignment of semantic mappings between the source and target domains, DHTL incorporates Canonical Correlation Analysis \cite{hardoon2004canonical} to enable the matching of semantic representations of co-occurrence data pairs layer by layer. Consequently, the method learns a common semantic subspace that allows the utilization of labeled source features for model development in a target domain.\\


Previous instance-based methods focus on offline or batch learning problems, which assume that all training instances from the target domain are available in advance. However, this assumption may not hold true in many real-world applications. Several online HTL methods are capable of addressing scenarios where the target data sequence is acquired incrementally in real-time, while the offline source instances are available at the start of the training process. Since the labeled target instances are often extremely limited at the start of training, it is particularly important to transfer knowledge from source domains in these scenarios. We introduce two online instance-based methods here.

\paragraph*{OHKT:}
\textbf{O}nline \textbf{H}eterogeneous \textbf{K}nowledge \textbf{T}ransition (OHKT) \cite{OHKT} 
bridges the target (image) and source (text) domains by generating pseudo labels for co-occurrence data, which consist of text-image pairs. The approach involves training a classifier on the labeled source data and using it to assign pseudo labels to the co-occurrence data. These pseudo-labels are subsequently utilized to assist the online learning task in the target domain, facilitating the transfer of knowledge from the source domain.

\paragraph*{OHTHE:} Directly using co-occurrence data can be simplistic and may not capture the underlying nuances of similarity. Addressing this, \textbf{O}nline \textbf{H}eterogeneous \textbf{T}ransfer learning by \textbf{H}edge \textbf{E}nsemble (OHTHE) \cite{OHTHE} introduces a measure of heterogeneous similarity between target and source instances using co-occurrence data. Specifically, OHTHE derives the similarity between target instance $x_{T,i}$ and source instance $x_{S,j}$ by incorporating co-occurrence pairs $\{(x^c_{S,k}, x^c_{T,k})\}^{n_C}_{k=1}$ into the similarity computation. The formulated heterogeneous similarity $S_{\text{heter}}$ is given by:
\begin{equation}
    S_{\text{heter}}(x_{T,i},x_{S,j})=\sum^{n_C}_{k=1}S_{\text{S}}(x_{S,j}, x^c_{S,k})S_{\text{T}}(x_{T,i}, x^c_{T,k}) ~,
\end{equation}
where $S_{\text{S}}$ and $S_{\text{T}}$ denote the similarity measures in the source and target domains, respectively. Notably, the Pearson correlation is employed as the similarity metric for both domains, ensuring consistency in the similarity evaluation. This similarity measure is then employed to guide the classification of unlabeled target instances by incorporating information from source labels. OHTHE achieves this by learning an offline decision function $h^{\text{off}}(X_T)$ for the target instances, accomplished through aligning the source label information for target instances using the similarity measure. Simultaneously, OHTHE utilizes target data to directly construct an online updated classifier $h^{\text{on}}(X_T) = W^{\top} X_T$. The final ensemble classifier is formed by combining $h^{\text{off}}(X_T)$ and $h^{\text{on}}(X_T)$ through a convex combination, and the method employs a hedge weighting strategy \cite{FREUND1997119} to update the parameters in an online manner.\\

In summary, this section has explored both offline \cite{TTL, HTLIC, DHTL} and online \cite{OHKT, OHTHE} instance-based methods. These methods are characterized by the use of an intermediate domain, with some \cite{DHTL,OHKT,OHTHE} employing a specific type of intermediate domain known as co-occurrence data. While some instance-based methods utilize traditional techniques such as matrix factorization \cite{TTL, HTLIC}, others incorporate deep neural networks \cite{DHTL}.


While instance-based methods are typically intuitive and effective for connecting heterogeneous source and target domains by leveraging supplementary data to discover underlying relationships, there are scenarios where obtaining an adequate amount of supplementary data is challenging. In such cases, instance-based methods may inadvertently lead to what is known as `over-adaptation'. Over-adaptation occurs when weakly correlated features, which lack semantic counterparts in the other domain, are compelled into a common feature space within the latent domain. This phenomenon can hinder the performance of transfer learning \cite{WOS:000500691600003}. Furthermore, there are situations in which acquiring intermediate data is not feasible due to various constraints. In such cases, it becomes imperative to explore alternative strategies that do not rely on the availability of intermediate domain data, such as feature representation-based methods.

\subsection{Feature Representation-based Method}
\label{feature-based}


In HTL, feature representation-based approaches hold a paramount position. These methods tackle the heterogeneity between the source feature space $\mathcal{X}_S$ and the target feature space $\mathcal{X}_T$ by aligning the heterogeneous spaces into a cohesive unified space, denoted as $\mathcal{X}$. This alignment is realized by learning two projection functions, as illustrated in
\begin{equation}
\label{eq:feature}
\begin{aligned}
\hat{x}_S &= \phi_S\left(x_S\right), &x_S &\in \mathcal{X}_S, &\hat{x}_S &\in \mathcal{X} ~, \\
\hat{x}_T &= \phi_T\left(x_T\right), &x_T &\in \mathcal{X}_T, &\hat{x}_T &\in \mathcal{X} ~,
\end{aligned}
\end{equation}
where $\phi_S\left(\cdot\right)$ and $\phi_T\left(\cdot\right)$ are the projection functions in the source and target domain, respectively.
In this unified space $\mathcal{X}$, the diverse features from the original heterogeneous spaces can be effectively compared and shared, paving the way for enhanced learning across different domains.

The primary goal of the feature representation-based method is to reduce the disparity between the source and target domains, with the evaluation of the similarity of their distributions being a critical initial step in this process.
In this context, the \textbf{M}aximum \textbf{M}ean \textbf{D}iscrepancy (MMD) \cite{gretton2006kernel} is employed as a measure of distribution similarity. MMD assesses the distances between the means of distributions in a Reproducing Kernel Hilbert Space (RKHS) according to the following formula:
\begin{equation}\label{MMD}
\text{MMD}({X}_S, {X}_T) = \left\| \frac{ 1}{n_S} \sum_{i = 1}^{n_S} \phi_S(x_{S,i}) -  \frac {1}{n_T} \sum_{i =1}^{n_T} \phi_T(x_{T,i})\right\| ~.
\end{equation}
Minimizing the MMD value implies a reduction in distribution disparity between the source and target domains, indicating that the features in both domains are becoming more similarly distributed. Achieving a minimized MMD value is pivotal as it signifies a successful alignment of feature distributions across two domains, which is a fundamental step toward mitigating the discrepancy between them. In addition to the MMD metric, there are other measures such as Soft-MMD \cite{STN} and the $\mathcal{A}$-distance \cite{kifer2004detecting}. However, these are not as commonly utilized as the MMD metric.


Feature representation-based methods are mainly divided into two fundamental operations: feature mapping and feature augmentation. 
The feature mapping operation involves projecting source and target features into a shared representation space. This mapping aims to align the feature distributions of two domains and mitigate the underlying heterogeneity, thus facilitating the seamless transfer of knowledge between them. On the other hand, feature augmentation methods incorporate both domain-invariant features and the original domain-specific features from each domain. This approach not only considers a common subspace for comparing heterogeneous data but also keeps the domain-specific patterns, leading to more comprehensive and effective feature representations.

\subsubsection{Feature Mapping}
\label{mapping}

Feature mapping refers to the process of transforming or encoding input features into new representations that are better suited for specific tasks or analysis. In the context of traditional feature mapping, the objective is to extract informative  features from the original data. This transformation can utilize various techniques depending on the nature of the data and the specific tasks involved. For example, Principal Component Analysis (PCA) \cite{PCA} is an unsupervised dimensionality reduction technique that aims to reduce the data dimensionality and retain the most informative features by maximizing its variance after transformation. With label information, Linear Discriminant Analysis (LDA) \cite{LDA} is a supervised dimensionality reduction technique. Its primary objective is to find a projection that not only reduces the dimensionality but also maximizes the distinction among different classes. By achieving this, LDA effectively transforms the data into a lower-dimensional space where class distinction is significantly improved.

To handle heterogeneity in the original feature spaces, feature mapping projects the original features of the source and target domains into an aligned feature space. This process seeks to extract valuable features from original data while capturing relevant information and harmonizing the distributions of both domains. Feature mapping techniques in HTL encompass various approaches, including linear transformations, nonlinear mappings, and more complex deep learning architectures.

\begin{figure}[htbp]
\centerline{\includegraphics[width=90mm]{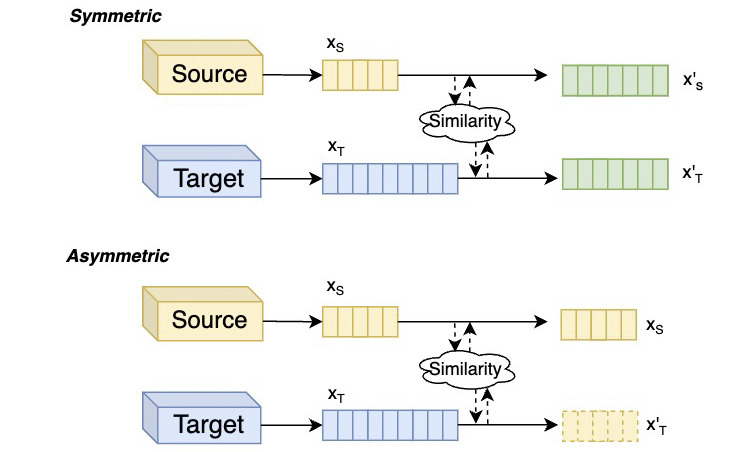}}
\caption{Two feature mapping methods: symmetric (upper) and asymmetric (lower). The asymmetric method depicts mapping from target to source dimensions (shown in the figure), providing an alternative approach to projecting source to target (not depicted in the figure).}

\label{fig3}
\end{figure}

As shown in Figure \ref{fig3}, the feature mapping approaches can be categorized into two types: symmetric transformation and asymmetric transformation. As illustrated in Eq. \ref{eq:feature}, the goal of symmetric feature mapping is to learn a pair of projections $\phi_S$ and $\phi_T$, which map the source domain data $x_S$ and the target domain instances $x_T$, respectively, into a shared feature space. 
In contrast, asymmetric feature mapping methods focus on learning a single projection function $\phi$. This function is used to map either the source features $x_S$ into the feature space of the target domain $x_T$ or vice versa. The ultimate goal of this approach is to find a transformation that adapts the features of one domain to those of the other, thereby minimizing the differences between $\phi\left(x_S\right)$ and $x_T$ or $\phi\left(x_T\right)$ and $x_S$.

We will first delve into asymmetric methods. One prominent example is the \textbf{I}nformation-\textbf{T}heoretic \textbf{M}etric \textbf{L}earning (ITML) method \cite{ITML}, which employs a linear transformation matrix $W$. This matrix facilitates the translation of target instances $x_T$ into the source domain through $W$ or conversely, morphs source instances $x_S$ into the target domain using $W^\top$. Despite its merits, ITML encounters constraints when the dimensionalities of both domains aren't equivalent, thereby confining it to homogeneous contexts. To address this limitation, the \textbf{A}symmetric \textbf{R}egularized \textbf{C}ross-domain \textbf{t}ransformation (ARC-t) method \cite{brain2011what} learns the transformations in kernel space. This innovation allows the method to be applied in more general cases where the domains do not have the same dimensionality. Following this idea, asymmetric feature mapping can convert instances from one domain into another heterogeneous domain, thereby transforming a heterogeneous transfer learning problem into a homogeneous one. 
Having established the foundational concepts in asymmetric feature mapping, the following paragraphs will delve deeper into specific examples to further elucidate these principles and demonstrate their practical applications.


\paragraph*{CDLS:} 
The \textbf{C}ross-\textbf{D}omain \textbf{L}andmark \textbf{S}election (CDLS) method \cite{CDLS} establishes a common homogeneous space by projecting the target data into a subspace using PCA. To bring the source-domain data into this subspace, CDLS utilizes a feature transformation matrix denoted as $A$ which helps to eliminate domain difference. By learning $A$, the technique aims to match the marginal distributions $P_T\left(X_T\right)$ and $P_S\left(A^{\top} X_S\right)$, while also aligning the conditional distributions $P_T\left(y_T \mid X_T\right)$ and $P_S\left(y_S \mid A^{\top} X_S\right)$. 

\paragraph*{SHDA-RF:} Utilizing information from label distributions, \textbf{S}upervised \textbf{H}eterogeneous \textbf{D}omain \textbf{A}daptation via \textbf{R}andom \textbf{F}orests (SHDA-RF) \cite{SHDA-RF} derives the pivots that serve as corresponding pairs, bridging the gap between the heterogeneous source and target domains.
The SHDA-RF process begins by identifying $N_p$ pivots from both the source and target random forest models, which share the same label distributions. These pivots act as connections between the heterogeneous feature spaces. Utilizing the $N_p$ derived pivots, the method estimates feature contribution matrices $W_S \in \mathbb{R}^{N_p \times d_S}$ and $W_T \in \mathbb{R}^{N_p \times d_T}$. Subsequently, a projection matrix $P_S$ is learned from these matrices, enabling the mapping of source features $X_S$ to target features $X_T$.

\paragraph*{SHFR:} 
Instead of relying on instance correspondences, \textbf{S}parse \textbf{H}eterogeneous \textbf{F}eature \textbf{R}epresentation (SHFR) method \cite{SHFR} learns the feature mapping function based on weight vectors $w_S$ and $w_T$, assuming linear classifiers. By maximizing $w_T^{ \top} W w_S$ or minimizing $\left\|w_T-W w_S\right\|$, SHFR learns a mapping function $W$ that can align source and target domains effectively. \\

While asymmetric feature mapping offers flexibility and ease of implementation with only one projection to learn \cite{brain2011what}, symmetric feature mapping is more commonly employed due to its versatility in HTL. Symmetric feature mapping involves the transformation of both feature domains into a shared latent feature space. Specifically, symmetric feature mapping transforms the heterogeneous features into one shared space. The mapping transformation could be as simple as 
\begin{equation}
\hat{X}_S=X_S P_S, \quad \hat{X}_T= X_T P_T ~,
\end{equation}
where $P_S \in \mathbb{R}^{d_S \times d} $ and $P_T \in \mathbb{R}^{d_T \times d}$ are projection matrices that map the source and target features into a common space $\mathbb{R}^d$. 
By learning a common representation, symmetric feature mapping facilitates better alignment of feature distributions and enhances the generalization capability of the model by capturing underlying structures that are relevant to both domains.
In the following paragraphs, we will explore various approaches and algorithms that utilize symmetric feature mapping to address HTL challenges.

\paragraph*{HeMap:} \textbf{He}terogeneous Spectral \textbf{Map}ping (HeMap) \cite{HeMap} learns two linear transformation matrices $P_S, P_T$ using spectral embedding in the following optimization objective,
\begin{equation}
\begin{aligned}
    \min _{P_S, P_T} &\|\hat{X}_T P_T- X_T\|^2+\|\hat{X}_S P_S- X_S\|^2+ \\
    &1/2 \cdot \beta \cdot \left( \|\hat{X}_S P_T- X_T\|^2+\|\hat{X}_T P_S, X_S\|^2 \right) ~,
\end{aligned}
\end{equation}
where $\hat{X}_S$ and $\hat{X}_T$ are projected source and target data.
The primary objective of this optimization is to enable the projection to enhance data similarity while preserving inherent structural characteristics. Preserving structural information is of paramount importance, particularly for accurate data classification \cite{huang2020heterogeneous}.

\paragraph*{DACoM:} The \textbf{D}omain \textbf{A}daptation by \textbf{Co}variance \textbf{M}atching (DACoM) \cite{DACoM} introduces transformations that incorporate the zero-mean characteristics into the mapped features. Specifically, it performs the following transformations to automatically make the two first moments equal: 
\begin{equation}
\begin{aligned}
\hat{x}_S&=\left(x_S-\bar{X}_S\right)P_S, &\hat{x}_S \in \mathbb{R}^d ~,\\ \hat{x}_T&=\left(x_T-\bar{X}_T\right)P_T, &\hat{x}_T \in \mathbb{R}^d ~,
\end{aligned}
\end{equation} 
where $\bar{X}_S$ and $\bar{X}_T$ denote the means of $X_S$ and $X_T$, respectively. By doing so, the first moments are automatically equal and DACoM minimizes the gap of their covariance matrices in both domains to learn more consistent distributions of the projected instances.

\paragraph*{DAMA:} 
Given multiple heterogeneous source domains, Heterogeneous \textbf{D}omain \textbf{A}daptation using \textbf{M}anifold \textbf{A}lignment (DAMA) \cite{HDAMA} considers each domain as a manifold, represented by a Laplacian matrix constructed from an affinity graph that captures relationships among instances. DAMA aims to reduce the dimensionality of feature space while preserving manifold topology through generalized eigenvalue decomposition. This process generates a lower dimensional feature space that can be utilized for transfer learning across domains. However, DAMA assumes that the data follows a manifold structure. 

\paragraph*{LPJT:} While geometric manifold structures are pivotal as discussed in previous methods, other latent factors also play a crucial role in establishing a connection between the source and target domains. Factors such as landmark instances, which are a select subset of labeled source instances closely distributed to the target domain, are of particular importance. \textbf{L}ocality \textbf{P}reserving \textbf{J}oint \textbf{T}ransfer (LPJT) method \cite{LPJT} proposes a unified objective to optimize all aspects at the same time. 
The transformation matrices are learned by minimizing the marginal and conditional MMD between the common space of source and target domains, reducing domain shifts while preserving local manifold structures through the minimization of intra-class instance distance and the maximization of inter-class instance distance. By doing so, the LPJT method establishes a connection between heterogeneous source and target domains. Additionally, the LPJT method incorporates a re-weighting strategy for landmark selection, which aids in selecting pivot instances as bridges for effective knowledge transfer.

\paragraph*{ICDM:} The \textbf{I}nformation \textbf{C}apturing and \textbf{D}istribution \textbf{M}atching (ICDM) method \cite{ICDM} introduces a similar approach to LPJT by utilizing MMD for aligning domain distributions but extends its scope beyond distribution matching. ICDM places emphasis on preserving original feature information through the minimization of reconstruction loss between original and reconstructed data. ICDM can capture and maintain the essential characteristics of original features during the domain adaptation process.\\

In HTL, a recurrent challenge is the scarcity of label information within the target domain. This sparsity underscores the paramount importance of effectively harnessing whatever limited labels are available in the target setting \cite{zhou2022label}. In response to this challenge, several methods have been formulated. Some methods use label information to enforce the similarity of projected data points in the same class across different domains. Others incorporate a supervised classification loss to the objective function. 

\paragraph*{CDSPP:} \textbf{C}ross-\textbf{D}omain \textbf{S}tructure \textbf{P}reserving \textbf{P}rojection (CDSPP) algorithm \cite{CDSPP} incorporates a symmetric feature mapping approach to enforce the proximity of the projected instances belonging to the same class, regardless of their original domains, using the similarity matrix of the training instances derived from the label information.

\paragraph*{STN:} \textbf{S}oft \textbf{T}ransfer \textbf{N}etwork (STN) \cite{STN} simultaneously learns a domain-shared subspace and a classifier $f(\cdot)$. 
The STN constructs two projection networks that are dedicated to mapping the data from both the source and target domains, $X_S$ and $X_T$, into $\hat{X}_S$ and $\hat{X}_T$ respectively, within a common domain-invariant subspace. The optimization process involves minimizing the classification loss $\mathcal{L}_{C}$ calculated over $n_S$ source instances and $n_T^l$ labeled target instances, together represented as $X_a=\left[\hat{X}_S, \hat{X}_{T}^{l}\right]$ and their corresponding labels $Y_a=\left[Y_S, Y_{T}^{l}\right]$. Additionally, a Soft Maximum Mean Discrepancy (Soft-MMD) loss is employed to align both the marginal and conditional distributions between the domains. 
The objective function of STN includes a classification loss and Soft-MMD loss together as:
\begin{align}
     \mathcal{L}=&\mathcal{L}_{C}\left[\mathrm{Y}_a, f\left(\mathrm{X}_a\right)\right] +\text{Soft-MMD}\left[\hat{X}_S,\hat{X}_T\right] ~.
\end{align}
The Soft-MMD is an extension of the MMD concept. The MMD mainly focuses on the divergence in marginal distributions. The Soft-MMD further accounts for the discrepancies in conditional distributions across different domains. The Soft-MMD is defined as, 
\begin{equation}\label{Soft-MMD}
    \text{Soft-MMD}\left[\hat{X}_S,\hat{X}_T \right] = \text{MMD}\left[\hat{X}_S,\hat{X}_T\right] + {Q_c} ~,
\end{equation} 

and
\begin{equation}\label{soft-MMD}
Q_c=\sum_{k=1}^C\left\|\frac{1}{n_S^k} \sum_{i=1}^{n_S^k} \hat{X}^{k, i}_S-\frac{\sum_{i=1}^{n_l^k} \hat{X}^{k, i}_l+\sum_{i=1}^{n_u} \alpha_i^{(r)} \hat{X}^i_u}{n_l^k+\sum_{i=1}^{n_u} \alpha_i^{(r)}}\right\|^2 ~.
\end{equation}
Here, $\alpha_i^{(r)}=\frac{r \times y^{k, i}_u}{R}$ denotes the adaptive coefficient with $R$ as the total number of iterations and $r$ indicating the current iteration. To address the scarcity of labeled target instances, Soft-MMD leverages the unlabeled target data $X_{T}^{u}$ and assigns $C$-dimensional soft labels $y_{T}^{u}=f(\hat{X}_{T}^{u})$, which represent the probabilities of the projected data $\hat{X}_{T}^{u}$ belonging to $C$ categories. This approach also introduces an adaptive coefficient that gradually increases the weight of the predicted labels. 

\paragraph*{SCT:} \textbf{S}emantic \textbf{C}orrelation \textbf{T}ransfer (SCT) \cite{SCT} aims to transfer knowledge of semantic correlations among categories from the source domain to the target domain. The method measures semantic correlations by cosine similarity between different local centroids in the source domain. To achieve this, SCT uses two projection functions to map source and target features into a shared space. The optimization process involves minimizing a loss function that encompasses several components: the discrepancy in marginal distribution, the discrepancy in conditional distribution, the discrepancy in cosine distances among classes across both domains and the supervised classification loss. Through this approach, SCT not only encourages the learning of domain-invariant features that reduce the mixing of features from different classes but also enhances the discriminative ability of categories within the target domain.\\




Many HTL methods focus on addressing either feature discrepancy or distribution divergence one at a time. However, optimizing one can enhance the other. Subsequently, some methods further optimize both of them simultaneously.

\paragraph*{HDAPA:} \textbf{H}eterogeneous \textbf{D}omain \textbf{A}daptation Through \\
\textbf{P}rogressive \textbf{A}lignment (HDAPA) \cite{HDAPA} simultaneously optimizes feature difference and distribution divergence. This method maps the domain features $X_S, X_T$ into new representations $S_S, S_T$ in a shared latent space, using two domain-specific projections $P_S, P_T$ and a common codebook $B$. It uses the MMD metric \eqref{MMD} to measure distribution divergence. By solving the variables $P, B, S$ alternatively using the objective function illustrated as follows, 
\begin{equation}
\min _{B, S} \underbrace{\underbrace{\mathcal{C}_1\left(X_S, X_T, P, B, S\right)}_{\text {feature alignment }}+\underbrace{\alpha \mathcal{C}_2\left(S_S, S_T\right)}_{\text {distribution alignment }}}_{\text{progressive alignment}}+\underbrace{\beta R(S_S, S_T)}_{\text {constraint }} ~.
\end{equation}
The algorithm progressively learns the new representations for source and target domains.

\paragraph*{HANDA:} Similarly,  \textbf{H}eterogeneous
\textbf{A}dversarial \textbf{N}eural \textbf{D}omain 
\textbf{A}daptation (HANDA) \cite{HANDA} conducts both feature and distribution alignment within a unified neural network architecture. The method achieves this by using a shared dictionary learning approach to project heterogeneous features into a common latent space, thereby handling heterogeneity while alleviating feature discrepancy. An adversarial kernel matching method is then employed to reduce distribution divergence. Finally, a shared classifier is used to minimize the shared classification loss. 
\\

Nevertheless, lower-order statistics do not always fully characterize the heterogeneity of the domains \cite{jin2022heterogeneous}. Some methods employ neural network based structures to map the heterogeneous feature domains to one shared representation space.

\paragraph*{TNT:} \textbf{T}ransfer \textbf{N}eural \textbf{T}rees (TNT) method \cite{TNT, WOS:000476797800013} jointly solves cross-domain feature mapping, adaptation, and classification in a neural network based architecture. TNT learns the source and target feature mapping $\phi_S$ and $\phi_T$ respectively and updates them to minimize the prediction error for the labeled source data $X_S$ and target domain data $X_L$. Due to the lack of label information for the unlabeled target-domain data $X_U$, the method preserves the prediction and structural consistency between $X_L$ and $X_U$ to learn $\phi_T$ with $X_U$.\\

In this subsection, we discussed feature mapping methods, which present sophisticated approaches to bridge the gap between source and target domains in HTL by projecting them into a shared, domain-invariant subspace. The feature mapping methods discussed can be categorized into asymmetric \cite{ITML, brain2011what, CDLS, SHDA-RF, SHFR} and symmetric transformations, with symmetric transformation being the predominant type. These methods aim to align the source and target domains by considering various factors that include domain distribution \cite{DACoM}, manifold structure \cite{HDAMA}, and landmark selection \cite{LPJT}. Given that target label information is often limited, some methods \cite{CDSPP, STN, SCT} effectively utilize it by employing classification loss or enforcing similarity among instances within the same category. Regarding projection methods, approaches vary from using basic matrices \cite{HeMap} and dictionary learning \cite{HDAPA, HANDA} to employing neural networks \cite{TNT}.




\subsubsection{Feature Augmentation} 
\label{augmentation}

Within feature-based methods in HTL, feature augmentation is another pivotal strategy to align the heterogeneous domains in a common subspace. Distinct from feature mapping methods, which predominantly search for domain-invariant representations, feature augmentation methods go a step further by incorporating domain-specific features. It augments the original domain-specific features with the domain-invariant features learned through transformations. By doing so, it not only learns a common subspace where the heterogeneous data can be compared but also keeps domain-specific patterns \cite{pan2010cross}.

Feature augmentation methods were first applied in homogeneous transfer learning. Consider source domain feature $x_S \in \mathbb{R}^d$ and target domain feature $x_T\in \mathbb{R}^d$, the features in source and target domains can be augmented to be $\left[x_S, x_S, \mathbf{0}\right]$ and $\left[x_T, \mathbf{0}, x_T\right]$ respectively \cite{daume2009frustratingly}, where $\mathbf{0} \in \mathbb{R}^d$ is a zero vector. In this way, the augmented feature has both domain-invariant and domain-specific spaces. However, in the context of HTL, direct concatenation of features becomes a challenge due to the dimensionality disparities between the domains. This necessitates a deeper dive into creating a common space for both domains. Consequently, the processes of heterogeneous feature augmentation become intertwined with heterogeneous feature mapping.

\begin{figure}[htbp]
\centerline{\includegraphics[width=90mm]{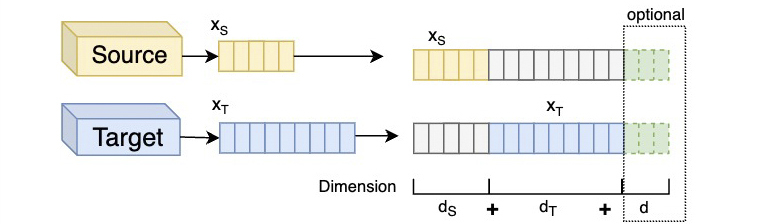}}
\caption{Feature augmentation method.} 


\label{fig_feature_agumentation}
\end{figure}
\paragraph*{SHFA:} In the \textbf{S}emi-supervised \textbf{H}eterogeneous \textbf{F}eature \\\textbf{A}ugmentation (SHFA) method \cite{HFA} \cite{SHFA}, source features $x_S \in \mathbb{R}^{d_S}$ and target features $x_T\in \mathbb{R}^{d_T}$are augmented as,  
\begin{equation}
\hat{x}_S=\left[x_S P_S,\space x_S,\space \mathbf{0}^{d_T}\right], \quad \hat{x}_T=\left[x_T P_T,\space \mathbf{0}^{d_S},\space x_T\right]~,
\end{equation}
where $P_S \in \mathbb{R}^{d_S \times d} $ and $P_T \in \mathbb{R}^{d_T \times d}$ are two projection matrices that map the source and target features into a shared common space $\mathbb{R}^d$;  $\mathbf{0}^{d_S}$ and $\mathbf{0}^{d_T}$ are zero vectors. By performing this feature augmentation, the heterogeneous source and target domains are effectively connected in a $\left(d + d_S + d_T \right)$-dimensional common space, enabling the transfer of knowledge and information between the two domains.

The alternative strategy discards the concept of common feature space. Instead, it initializes the source and target features as, 
\begin{equation}
\hat{x}_S=\left[x_S, \mathbf{0}^{d_T}\right], 
\hat{x}_T=\left[\mathbf{0}^{d_S}, x_T\right]~,
\end{equation}
which reduces the dimensionality from $d+d_S+d_T$ to $d_S+d_T$. This reduction can yield advantages in computational efficiency.

\paragraph*{DCA \& KPDA:} 
\textbf{D}iscriminative \textbf{C}orrelation \textbf{A}nalysis (DCA) \cite{DCA} and \textbf{K}nowledge \textbf{P}reserving and \textbf{D}istribution \textbf{A}lignment (KPDA) \cite{KPDA} augment the target features as $\hat{x}_T=\left[x_T P, x_T\right]$ where $P$ is a learnable matrix, which can avoid the problem of the curse of dimensionality in SHFA.

\paragraph*{Sym-GANs:} Equipped with deep learning techniques, \cite{SymGAN} proposes \textbf{Sym}metric \textbf{G}enerative \textbf{A}dversarial \textbf{N}etwork\textbf{s} (Sym-GANs) algorithm. This algorithm trains one Generative Adversarial Network (GAN) to map the source features to target features and another GAN for reverse mapping. Using labeled source domain data $x_S$ and target domain data $x_T$, the Sym-GANs algorithm learns bidirectional mappings denoted by $\mathcal{G}_T:x_S \rightarrow x_T$ and $\mathcal{G}_S:x_T \rightarrow x_S$. With these mappings, augmented features can be obtained: 
\begin{equation}
\begin{aligned}
\hat{x}_S &= \left [ \mathcal{G}_S(\mathcal{G}_T(x_S)); \mathcal{G}_T(x_S) \right ] \in \mathbb{R}^{d_S+d_T} ~, \\
\hat{x}_T &= \left [ \mathcal{G}_S(x_T); \mathcal{G}_T(\mathcal{G}_S(x_T))\right ] \in \mathbb{R}^{d_S+d_T} ~.
\end{aligned}
\end{equation}
These newly generated representations are then used for training a classifier of target instances for enhanced discriminative capability.\\

Some methods assume that instances from both the source and target domains share identical feature spaces. As a result, they construct a unified instance-feature matrix that includes all instances across both domains. By addressing the matrix completion challenge and subsequently reconstructing the ``ground-truth'' feature-instance matrix, they obtain enhanced features within this common space.

\paragraph*{HTLA:} Given a set of  $n_S^l$ labeled instances $\{(X_S^l, y_S^l)\}$ from source domain, $n_S^u$ unlabeled instances $\{X_S^u\}$ from source domain, $n_T^u$ unlabeled instances $\{X_T^u\}$ from target domain, and corresponding pairs $\{(X_S^c, X_T^c)\}$ between the source and target domains, \textbf{H}eterogeneous \textbf{T}ransfer \textbf{L}earning through \textbf{A}ctive correspondences construction (HTLA) method \cite{HTLA} first builds a unified instance-feature matrix for all the instances. To address missing data, zero-padding is employed, leading to the matrix $\mathbf{X}$, defined as,
\begin{equation}
\textbf{X} = \left[
\begin{matrix}
{X_S^l} & {\mathbf{0}^{n_S^l,d_T}} \\
{X_S^u} & {\mathbf{0}^{n_S^u,d_T}} \\
{X_S^c} & {X_T^c} \\
{\mathbf{0}_{n_T^u,d_S}} & {X_T^u}  
\end{matrix}\right] ~.
\end{equation}
Subsequently, the missing entries within $\mathbf{X}$ undergo a recovery procedure accomplished through a matrix completion mechanism that is based on distribution matching, particularly utilizing the MMD. The final result is the fully recovered and completed matrix $\mathbf{X}$. A singular value decomposition is then applied to $\mathbf{X}$,
resulting in the projection of domain data into a shared latent space defined by the top $r$ singular vectors, expressed as $X=\mathbf{U}_r \boldsymbol{\Sigma}_r \mathbf{V}_r$. This projection yields the transformed feature matrix $\mathbf{Z}=X \mathbf{V}_r$. HTLA trains a classifier on the new feature representations of the source domain labeled data, comprising the first $\left(n_S^l+n_S^u\right)$ rows of $\mathbf{Z}$, and applies it to predict on the target domain data, encompassing the last $n_T^u$ rows of $\mathbf{Z}$.

\paragraph*{MKL:} 
Corresponding pairs employed in HTLA can be missing in some situations and \textbf{M}ultiple \textbf{K}ernel \textbf{L}earning (MKL) \cite{NMF} are proposed to address this problem. Given $n_S$ labeled source domain data $X_S$ and $n_T$ target domain data $X_T$, including a few labeled instances and unlabeled ones, MKL augments the data using zero padding as follows,
\begin{equation}\label{zero matrix}
\textbf{X} = \left[
\begin{matrix}
{X_S} & {\mathbf{0}^{n_S,d_T}} \\
{\mathbf{0}^{n_T,d_S}} & {X_T} 
\end{matrix}\right] ~.
\end{equation}
The approach introduces two latent factor matrices: $\mathbf{U} \in \mathbb{R}^{\left(n_S+n_T\right) \times k}$, which serves as the latent representations for matrix $X$, and $\mathbf{V} \in \mathbb{R}^{\left(d_S+d_T\right) \times k}$, which acts as the dictionary for matrix completion. This framework facilitates the matrix completion process, leading to the acquisition of a latent feature representation, denoted as $\hat{X} = \mathbf{U}\mathbf{V}^T$. 

\paragraph*{Deep-MCA:} Different from previous methods that rely on conventional matrix completion techniques, \textbf{Deep} \textbf{M}atrix \textbf{C}ompletion with \textbf{A}dversarial Kernel Embedding (Deep-MCA) \cite{DeepMCA} proposes a deep learning based framework. This approach employs an auto-encoder architecture denoted as $X_r = V(W(\textbf{X}))$ to perform matrix completion on the augmented matrix defined in Eq. \eqref{zero matrix} above, where $V(\cdot)$ represents decoder and $W(\cdot)$ represents encoder. By applying the encoder $W$ to the augmented features $\left[{X_S}, {\mathbf{0}^{n_S \times d_T}}\right]$ and $\left[{\mathbf{0}^{n_T \times d_S}}, {X_T}\right]$, mapping them into a Reproducing Kernel Hilbert Space, the method can use the newly generated representations to train a classifier for the target domain.\\

In this subsection, we discussed feature augmentation methods, which focus on enriching the domain-invariant feature space while preserving domain-specific features. Various techniques are employed to achieve this. Some methods utilize projection matrices \cite{SHFA, DCA, KPDA} or neural networks \cite{SymGAN}, drawing on approaches similar to feature mapping, to construct the domain-invariant space. Additionally, a particularly prevalent and effective method known as matrix completion is often used to augment the feature space in heterogeneous domain scenarios \cite{HTLA, NMF}.

\hfill
\\
For data-based methods, we delve into their intricacies, providing a comprehensive examination of their workings and nuances. While the effectiveness of data-based methods is well-documented, they do have limitations. Their primary drawback is the prerequisite for extensive training data from at least one of the domains, combined with the demand for substantial computational resources for parameter learning. This can pose challenges in scenarios with restricted data availability. Furthermore, the dependence on incorporating source data can raise significant data privacy concerns, especially when handling sensitive or proprietary information, thereby limiting the applicability of these methods in various domains.
To tackle these challenges, the paradigm of transferring well-developed models from the source domains offers an attractive alternative. We explore this avenue further in the subsequent section on model-based methods.


\section{Model-based Method}
\label{model-based}


Model-based methods in HTL primarily focus on transferring a source domain's model structure and parameters to a target domain. Specifically, given source data $X_S$ and source labels $y_S$, a source model is initially trained to obtain the optimal parameters $W_S$. Subsequently, these parameters $W_S$ guide the formulation of the parameters in the target model $W_T$. 

Two primary strategies are employed to leverage $W_S$ to influence $W_T$: parameter regularization and parameter tuning. Parameter regularization methods involve learning target models with a regularization term $\left\| W_S-W_T \right\|$. The target model's parameters $W_T$ start with random initialization and are adjusted to align with the characteristics of the target domain, while being regularized to prevent significant deviation from $W_S$. In contrast, parameter tuning initially sets the parameters $W_T$ to be equal to $W_S$ and subsequently adapts them to the target domains through fine-tuning. This strategy ensures that the target model parameters are initially aligned with those of the source model, and are later refined to accommodate the distinct characteristics of the target datasets.



\subsection{Parameter Regularization Method}
\label{regularization}

\begin{figure}[htbp]
\centerline{\includegraphics[width=90mm]{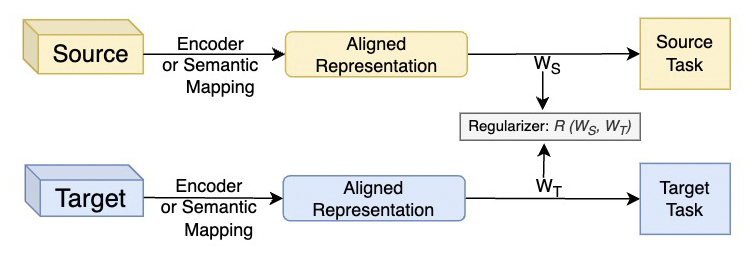}}
\caption{Parameter regularization method.}
\label{fig4}
\end{figure}

Parameter regularization methods, as shown in Fig. \ref{fig4}, aim to bridge the gap between the parameters of source and target models by introducing regularizers on their parameters. These techniques serve a dual purpose. First, they encourage the target models to embrace similar parameter values as those of the source models, thereby enabling them to leverage the general knowledge and patterns acquired from the source domain. Second, these methods provide the target models the flexibility needed to adapt to the distinct characteristics of the target domain. This adaptability is instrumental in enhancing the accuracy of the target model and safeguarding against over-fitting, a common concern when dealing with limited data from the target domain.                     

However, it is worth noting that the widely existing difference between source and target feature spaces presents unique challenges, especially in scenarios involving multiple modalities. The model parameters learned in one domain may not be directly applicable to another domain due to variations in feature spaces. This disparity necessitates alignment processes to ensure the effective transfer of knowledge between source and target domains.

\paragraph*{REFORM:} The \textbf{RE}cti\textbf{F}y via heter\textbf{O}geneous p\textbf{R}edictor \textbf{M}apping (REFORM) \cite{REFORM} employs a semantic mapping to handle heterogeneity in either the feature or label space. 
By applying the semantic map  $\mathcal{M}$, a source model's parameters $\hat{W}_0$ are transformed to provide biased regularization that reflects prior knowledge for the target task's parameters $W \in \mathbb{R}^{d\times C}$ as in 
\begin{equation}
\label{REFORM}
    \min _W \frac{1}{N} \sum_{i=1}^N \ell\left(f\left(\mathbf{x}_i\right)-\mathbf{y}_i\right)+\lambda\left\|W-\mathcal{M}\left(\hat{W}_0\right)\right\|_F^2 ~.
\end{equation}
The REFORM deduces the semantic map $\mathcal{M}$ by learning a transformation matrix $T \in \mathbb{R}^{d\times d'}$. This matrix transforms the representation $\hat{W}_0 \in \mathbb{R}^{d'\times C}$ into $\mathcal{M}\left(\hat{W}_0\right)=T \hat{W}_0$ for the heterogeneous feature space. Similarly, REFORM can accommodate a heterogeneous label space by modifying $\mathcal{M}\left(\hat{W}_0\right)$.

\paragraph*{DTNs:} Weakly-shared \textbf{D}eep \textbf{T}ransfer \textbf{N}etwork\textbf{s} method (DTNs) \cite{DTNs} employs two $L_1$-layer stacked auto-encoders to derive aligned hidden representations from two heterogeneous domains. These aligned representations subsequently serve as input for the next sequence of $L_2$-layer models specific to each domain. Rather than directly enforcing parameter sharing across domains, DTNs opt for separate series of layers structured as follows,
\begin{align}
\nonumber X_S^{(l)} := h_S^{(l)}\left(X_S\right)&=s_e^s\left(\mathbf{W}_S^{(l)} X_S^{(l-1)}+\mathbf{b}_S^{(l)}\right) ~, \\
X_T^{(l)} := h_T^{(l)}\left(X_T\right)&=s_e^t\left(\mathbf{W}_T^{(l)} X_T^{(l-1)}+\mathbf{b}_T^{(l)}\right) ~,
\end{align}
where $s_e^s$ and $s_e^t$ are the encoders in source and target domains respectively, and $h_S^{(l)}\left(\cdot\right)$ and $h_T^{(l)}\left(\cdot\right)$ denote the $l$-th layer hidden representation in source and target domain respectively.
Under the assumption of weak parameter sharing, this method introduces a regularizer that governs the differences only between the parameters of the last few layers as
\begin{equation}
\Omega=\sum^{L_1+L_2}_{l=L_1+1}\left(\left\|\mathbf{W}_S^{(l)}-\mathbf{W}_T^{(l)}\right\|_F^2+\left\|\mathbf{b}_S^{(l)}-\mathbf{b}_T^{(l)}\right\|_2^2\right) ~.
\end{equation}
This design choice allows the initial $L_1$ layers to learn domain-specific features,  while the concluding $L_2$ layers specialize in identifying sharable knowledge across domains.. 

Parameter regularization methods, while effective in specific scenarios, can become time-consuming, particularly when dealing with significant domain shifts between  the source and target domains. This is because they rely on random initialization, which can hinder their effectiveness in adapting to domain-specific patterns. To address these challenges, parameter tuning methods have been introduced as an alternative solution.

\subsection{Parameter Tuning Method}
\label{tuning}

 



\begin{figure}[htbp]
\centerline{\includegraphics[width=85mm]{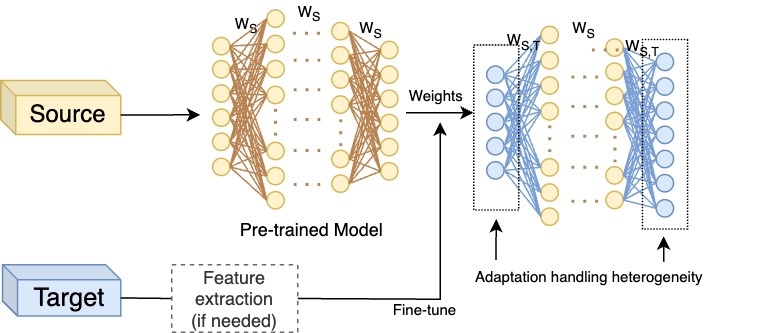}}

\caption{Parameter tuning method.}
\label{fig5}
\end{figure}

Parameter tuning methods in HTL, illustrated in Fig. \ref{fig5}, are designed to enhance the abilities of pre-trained models to perform tasks for which they have not been extensively trained. The goal is to adeptly tune the parameters of these models, enabling their adaptation and specialization for various downstream tasks across different domains.
Parameter tuning methods encompass two distinct phases: pre-training and fine-tuning.
In the pre-training phase, a model is trained on extensive, diverse datasets for general tasks, often broader in scope than the specific target tasks. This enables the model to capture general patterns, providing valuable insights applicable to a variety of downstream tasks.
In the subsequent fine-tuning phase, the pre-trained model's parameters are fine-tuned on smaller, task-specific target datasets. This process tailors the encoded features to the particular task. By leveraging knowledge from pre-training, the final model can potentially outperform one trained from scratch, especially when target labeled data is limited, on the target task.

One distinctive advantage of parameter tuning methods, which sets them apart from parameter regularization methods, is their effectiveness in reducing computational demands. By leveraging pre-trained models, these methods gain a strategic upper hand by initializing optimization processes from advantageous positions within the optimization landscape. This leads to faster convergence compared to starting the optimization process from random initial points.

The parameter tuning methods have proven highly effective across various domains, notably illustrated by the widespread application of models pre-trained on ImageNet \cite{huh2016makes, kornblith2019better} in the field of CV, and the utilization of BERT \cite{BERT} 
in NLP tasks. These examples underscore the versatility and efficacy of parameter tuning methods in diverse applications, details of which will be explored in the following sections.

\subsubsection{Pre-training in NLP}

In the NLP area, pre-training aims to learn patterns and probabilistic relationships from large amounts of text data. The primary objective of pre-trained language models is to estimate the likelihood of a sequence of words as in Eq. (\ref{eq:brace1}) or the likelihood of a sequence of words based on the context of the preceding words as in Eq. (\ref{eq:brace2}) 
\begin{empheq} [left={P\left(w\right)=\empheqlbrace}]{alignat=2}
    &P\left(w_1, w_2, \dots, w_n\right)
    \label{eq:brace1} \\
    &P\left(w_i, \dots, w_n \mid w_1, \dots, w_{i-1}\right) \label{eq:brace2}
\end{empheq}
where $w_i$ denotes the $i$-th word in one sentence. These models tackle a broad spectrum of NLP tasks, including but not limited to text prediction, text generation, text completion, and language understanding.


Among language models that have been pre-trained, transformer-based models have recently emerged as the most dominant. The transformer model, a neural network architecture introduced by \cite{vaswani2017attention}, is grounded in the concept of a multi-head self-attention mechanism. This mechanism allows the model to capture global dependencies and relationships among words in a sequence. Stemming from the transformer, two typical model structures have been developed: autoencoding models and autoregressive models. 

Autoencoding models aims to to learn compact representations of the input text in an unsupervised manner, typically designed for dimensionality reduction and feature learning. An autoencoder achieves this through two primary components: the encoder and the decoder. The encoder compresses the input data into lower-dimensional latent representations, and the decoder attempts to reconstruct the original input data from these compressed representations. The most renowned autoencoding model is BERT, which employs Masked Language Modeling to learn contextualized word representations. A percentage of the input tokens are randomly masked. 
The model is then trained to predict these masked tokens based on their surrounding context. This bidirectional training allows BERT to capture both the left and right context of a word, enabling it to learn deep contextual representations. Furthermore, during pre-training, BERT utilizes Next Sentence Prediction to understand the relationships between sentences by providing pairs of sentences to the model and training it to predict whether the second sentence logically follows the first sentence in the original text. This task helps BERT learn sentence-level representations and capture discourse-level information.


Autoregressive models adopt a decoder-only structure to model the conditional probability distribution of the succeeding token given the previous tokens in the sequence. These models are typically designed for text generation, dialogue generation, and machine translation. A key characteristic of autoregressive models is their dependence on previously generated tokens to inform the generation of subsequent tokens. During the pre-training process, the model predicts the next word or token in a sequence based on the preceding words or tokens. This sequential nature allows them to capture contextual information, thereby producing coherent and contextually relevant text. Notable autoregressive language models include the GPT series \cite{GPT-1, GPT-2, GPT-3, GPT-4}. 
Recently, ChatGPT, building upon the foundation of GPT-3.5, has emerged as a noteworthy advancement in the field of pre-trained models. Its success stems from the incorporation of reinforcement learning utilizing human feedback, a methodology that iteratively refines the model's alignment with user intent. By integrating capabilities from GPT-4, a significant multimodal model capable of processing both image and text inputs, ChatGPT evolves into a versatile problem-solving tool, proficient in producing text-based outputs for a wide array of tasks. The great triumph of ChatGPT, akin to others, can be primarily attributed to the use of a vast and diverse corpus of data, which encompasses various forms and tasks for extensive pre-training to obtain extremely large models. This comprehensive training empowers it to adeptly comprehend and generate language \cite{zhou2023comprehensive, wu2023brief}.

\subsubsection{Pre-training in CV}

In the CV area, pre-training has emerged as a strategy to address challenges posed by limited labeled data and complex visual tasks, capturing low-level visual features, such as edges, textures, and colors, from a vast amount of source data. Through learning these visual representations, pre-trained models can discern essential visual cues and patterns. Subsequently, these pre-trained models serve as a foundational starting point for more specific CV tasks, including image classification and object recognition tasks. 

Pre-training in CV has proven particularly valuable in scenarios where domain-specific data is either scarce or expensive to obtain. Models pre-trained on generic datasets, such as ImageNet, have exhibited consistent improvements when adapting to various domain-specific CV tasks \cite{huh2016makes, kornblith2019better, raghu2019transfusion}. For instance, in medical imaging, acquiring labeled data often requires expert annotations and incurs significant costs. Utilizing models pre-trained on general datasets substantially boosts model performance without requiring extensive labeled medical data \cite{raghu2019transfusion}. 

Another advantage of pre-trained models in CV is their ability to expedite the training process. Initializing a model with parameters from pre-trained models, instead of random initialization, can promote faster convergence and better local minima during the optimization process. This is particularly beneficial when working with large-scale image datasets, where training a deep network from scratch might be computationally prohibitive.


\subsubsection{Fine tuning}
Upon completing the pre-training phase, models enter the fine-tuned process, adapting to specific downstream tasks. The fine-tuning process enables the pre-trained models to adapt their learned representations to target domains, thereby enhancing their performance on particular tasks. Various strategies have emerged to navigate the fine-tuning process, including using smaller learning rates, applying reduced learning rates to initial layers, strategically freezing and then gradually unfreezing layers, or exclusively reinitializing the final layer. In scenarios where a pronounced disparity exists between the source pre-training tasks and the target application, extensive fine-tuning of the entire network may become requisite. 
These fine-tuning methodologies can be classified based on criteria such as which layers are modified and the amount of task-specific data leveraged. Subsequent sections will discuss two key categories in these fine-tuning techniques:

\paragraph*{Full versus Partial fine-tuning:} Fine-tuning methods, when distinguished based on the layers subjected to modification, fall into two categories: full and partial fine-tuning. Full fine-tuning necessitates that every layer of the pre-trained models be further trained using task-specific data. This comprehensive adjustment enables the model to tailor its parameters to the specificities of the target domain. \cite{simonyan2014very} shows that, for the localization task in the ImageNet Large Scale Visual Recognition Challenge \cite{ILSVRC15},  fine-tuning all layers outperforms tuning only the fully connected layers. However, as indicated in \cite{shen2021partial}, direct knowledge transfer from source data might not always be optimal due to potential biases or even negative influences on the target class in certain scenarios. In such instances, partial fine-tuning methods could provide a viable alternative. In partial fine-tuning methods, only a subset of layers within the pre-trained models is modified while the rest remain frozen, preserving their pre-trained knowledge and ensuring the retention of general features and representations. 
Partial fine-tuning proves particularly valuable when dealing with smaller task-specific datasets, mitigating overfitting risks and leveraging pre-existing knowledge. Notably, while the common approach leans toward fine-tuning the final layers, studies \cite{lee2022surgical} have underscored the occasional benefit of tuning initial or middle layers for certain tasks. Despite the considerable advantages of utilizing pre-trained models, their local fine-tuning can be computationally intensive and challenging. To address this issue, Offsite-Tuning \cite{xiao2023offsite} has been proposed, offering a privacy-preserving and efficient transfer learning framework. In this approach, the first and final layers of the pre-trained model function as an adapter, with the remaining layers compressed into an entity referred to as an emulator. This structure enables the fine-tuning of the adapter using the target data, guided by the static emulator. Subsequently, the fine-tuned adapter is plugged into the original full pre-trained model, enhancing its performance on specified tasks. Besides computational challenges, fine-tuning can reduce robustness to distribution shifts. Robust fine-tuning might be achieved by linearly interpolating between the weights of the original zero-shot and fine-tuned models ~\cite{wortsman2022robust}. Averaging the weights of multiple fine-tuned models with different hyperparameter configurations was shown to improve accuracy without increasing inference time \cite{wortsman2022model}.


\subsubsection{Handling Heterogeneity of Feature Spaces}

Adapting pre-trained models to specialized target datasets introduces challenges, particularly in reconciling heterogeneity in input dimensions between the pre-trained model and the target data.

In the field of NLP, early research utilized feature transfer approaches in pre-training methods, focusing on integrating learned feature representations, such as word embeddings, into target tasks. These endeavors aimed to capture semantic information from extensive source datasets and transfer knowledge to target domains with limited resources. But it is important to highlight that word embeddings may display heterogeneity across diverse datasets, due to various factors such as data sources, languages, or contexts. 

With the advent of transformer-based models in 2017, pre-training in NLP has shifted its focus toward parameter transfer methods. Unlike their predecessors, parameter transfer methods assume that the source and target domains share common model structures, parameters, or prior distributions of hyperparameters. Instead of transferring features produced by previous encoders as in feature transfer, the parameter transfer directly shares the model structure and parameters of the pre-trained models. 
By implicitly encoding semantic information into the model parameters, these models eliminate the need for the separate word embedding step inherent in previous feature transfer approaches. Instead, the input is represented as a collection of words or tokens in the language, addressing the heterogeneity in feature spaces across different domains. This innovative approach ensures that representations in varied domains are inherently homogeneous, thereby effectively handling the discrepancies in feature spaces without the necessity for explicit preprocessing. 

In the field of CV, addressing heterogeneity in feature spaces during pre-training can be challenging, especially when interfacing with datasets with varying image sizes, resolutions, or modalities. Simple data preprocessing often include actions such as resizing or cropping images to a fixed size, converting images to a standard color space, or normalizing pixel values \cite{noroozi2016unsupervised, rebuffi2017learning, kolesnikov2020big, talebi2021learning}. An alternative technique is feature extraction, which transforms images using a feature extractor to align with the input size of the pre-trained model. 
For example, ProteinChat \cite{guo2023proteinchat} uses a projection layer as a feature extractor, enabling a smooth and effective connection between the protein images and the subsequent pre-trained large language model. 

Another example is the Vision Transformer (ViT) \cite{dosovitskiy2020image}, which was inspired by the natural capability of using ``tokens'' to handle heterogeneity in NLP. 
ViT treats images as sequences of flattened patches, where each patch is linearly embedded and then processed by the transformer architecture. The transformer can efficiently capture long-range dependencies across patches using self-attention mechanisms. ViT also incorporates positional embeddings to preserve the spatial context, which gets lost amidst the patch-based transformation. Upon being pre-trained on large, diverse datasets, ViT can extract meaningful and universal features, thereby demonstrating adeptness at dealing with heterogeneity. Its inherent design facilitates bridging disparities between diverse datasets by comprehending both local and global image features, eliminating the necessity for explicit spatial operations, and thus maintaining homogeneity in feature spaces. 

Another interesting example is the Visual-Linguistic BERT \cite{vl-bert}, which further develops a unified architecture based on transformers to craft pre-trainable generic representations suited for visual-linguistic tasks. This model is capable of accepting both visual and linguistic embedded features as input. Each element of the input constitutes either a word from the input sentence or a region-of-interest from the input image. While their content features are domain-specific, the representation generated through multiple multi-modal transformer attention modules, is proficient in aggregating and aligning visual-linguistic information.

\section{Application Scenarios}

\label{application}
\begin{table*}[htbp]
\center
\caption{The summary of application scenarios. }
\begin{tabular}{cc}
\toprule
\textbf{Application}        & \textbf{Reference}  \\ \midrule
NLP                & \cite{wu2013cotransfer, HDAPA, STN, KPDA, HTLA, NMF, OHKT, CDLS, OHTHE, SHFR, CDSPP, DACoM, CWAN, MGLE, OTLMS, SHFA, ruder2019transfer, SGW, WOS:000429325500003, WOS:000631201900035, WOS:000707695400001, amini2009learning}              \\ 
CV                 & \cite{LPJT, CDLS, HDAPA, STN, DCA, NMF, DeepMCA, ICDM, CDSPP, CWAN, SymGAN, DHTL, WOS:000821313900005, WOS:000356863900003, WOS:000429325500003, WOS:000707695400001, WOS:000375360900012, jin2022heterogeneous, WOS:000564647800001, SCP, WOS:000576289700036, WOS:000462364200010, WOS:000595969000001, WOS:000356863900003}    \\
Biomedicine        & \cite{ye2019transfer, wang2022survmaximin2, guo2023proteinchat, liang2023drugchat, liang2023xraychat, HDAMA, DACoM, ji2023prediction}          \\ 
Multimodality      & \cite{TTL, OHTHE, DTNs, STN, HDAPA, DeepMCA, OHKT, ICDM, CDSPP, SCT, CWAN, DHTL}    \\ \bottomrule
\end{tabular} 
\label{table:application}
\end{table*}

\begin{table*}[htbp]
\begin{minipage}{\linewidth}
\centering
\caption{The summary of benchmark datasets.}
\begin{tabular}{P{4.5cm}cP{4cm}P{5.5cm}}
\toprule
\textbf{Dataset}    & \textbf{Year}   & \textbf{Task}  & \textbf{Method} \\ \midrule
\multirow{2}{*}{20 Newsgroups \footnote{\url{http://qwone.com/~jason/20Newsgroups/}}} & \multirow{2}{*}{1995} & \multirow{2}{*}{\shortstack{Text Classification,\\ Topic Modeling}} & \multirow{2}{*}{\cite{SHDA-RF, MGLE, OTLMS}} \\
& & & \\ 
\midrule
{\multirow{2}{*}{Multi-Domain Sentiment \footnote{\url{https://www.cs.jhu.edu/~mdredze/datasets/sentiment/}}}}
 & {\multirow{2}{*}{2007}} & {\multirow{2}{*}{\shortstack{Sentiment Analysis,\\ Text Classification}}} & {\multirow{2}{*}{\cite{MGLE, SHFA, KPDA, HTLA, SHFR}}}  \\
& & &   \\ \midrule
{\multirow{2}{*}{Cross-Lingual Sentiment \footnote{\url{https://zenodo.org/record/3251672}}}} & {\multirow{2}{*}{2010}} & {\multirow{2}{*}{\shortstack{Cross-Lingual\\ Sentiment Analysis}}} & {\multirow{2}{*}{\cite{SHFA, KPDA, HTLA, SHFR}}}   \\
& & &  \\ \midrule
{\multirow{2}{*}{Office \footnote{\url{https://faculty.cc.gatech.edu/~judy/domainadapt/}} + Caltech \footnote{\url{https://www.vision.caltech.edu/datasets/}}}} & {\multirow{2}{*}{2010}} & {\multirow{2}{*}{\shortstack{Object Recognition,\\ Image Classification}}} & {\multirow{2}{*}{\cite{WOS:000375360900012, LPJT, CDLS, HDAPA, STN, DCA, NMF, DeepMCA, ICDM, CDSPP}}} \\ & & &  \\ \midrule
{\multirow{2}{*}{Multilingual Reuters Collection \footnote{\url{https://archive.ics.uci.edu/dataset/259/reuters+rcv1+rcv2+multilingual+multiview+text+categorization+test+collection}}}}
 & {\multirow{2}{*}{2013}} & {\multirow{2}{*}{\shortstack{Multilingual Classification, \\ Sentiment Analysis}}} & \cite{SCT, WOS:000429325500003, WOS:000631201900035, WOS:000707695400001, SGW, HDAPA, STN, KPDA, HTLA, NMF, OHKT, CDLS, OHTHE, SHFR, CDSPP, DACoM, CWAN, MGLE, OTLMS}   \\ \midrule
{\multirow{2}{*}{NUS-WIDE \footnote{\url{https://lms.comp.nus.edu.sg/wp-content/uploads/2019/research/nuswide/NUS-WIDE.html}} + ImageNet \footnote{\url{https://www.image-net.org/}}}} & {\multirow{2}{*}{2015}} & {\multirow{2}{*}{\shortstack{Image Classification}}} & {\multirow{2}{*}{\cite{STN, HDAPA, DeepMCA, OHKT, ICDM, CDSPP, SCT, CWAN, DHTL, TNT}}} \\ & & &  \\ \midrule
{\multirow{2}{*}{Office-Home \footnote{\url{https://www.hemanthdv.org/officeHomeDataset.html}}}} & {\multirow{2}{*}{2017}} & {\multirow{2}{*}{\shortstack{Object Recognition,\\ Image Classification}}} & {\multirow{2}{*}{\cite{CDSPP, CWAN, LPJT}}} \\ & & &  \\ \midrule
{\multirow{2}{*}{Multilingual Amazon Reviews \footnote{\url{https://registry.opendata.aws/amazon-reviews-ml/}}}} & {\multirow{2}{*}{2020}} & {\multirow{2}{*}{\shortstack{Multilingual Sentiment Analysis,\\ Text Classification}}}  & {\multirow{2}{*}{\cite{KPDA, MGLE}}}          \\ & & &  \\ \bottomrule
\end{tabular}  
\label{table:dataset}
\end{minipage}
\end{table*}

\begin{figure}[htbp]
\centerline{\includegraphics[width=90mm]{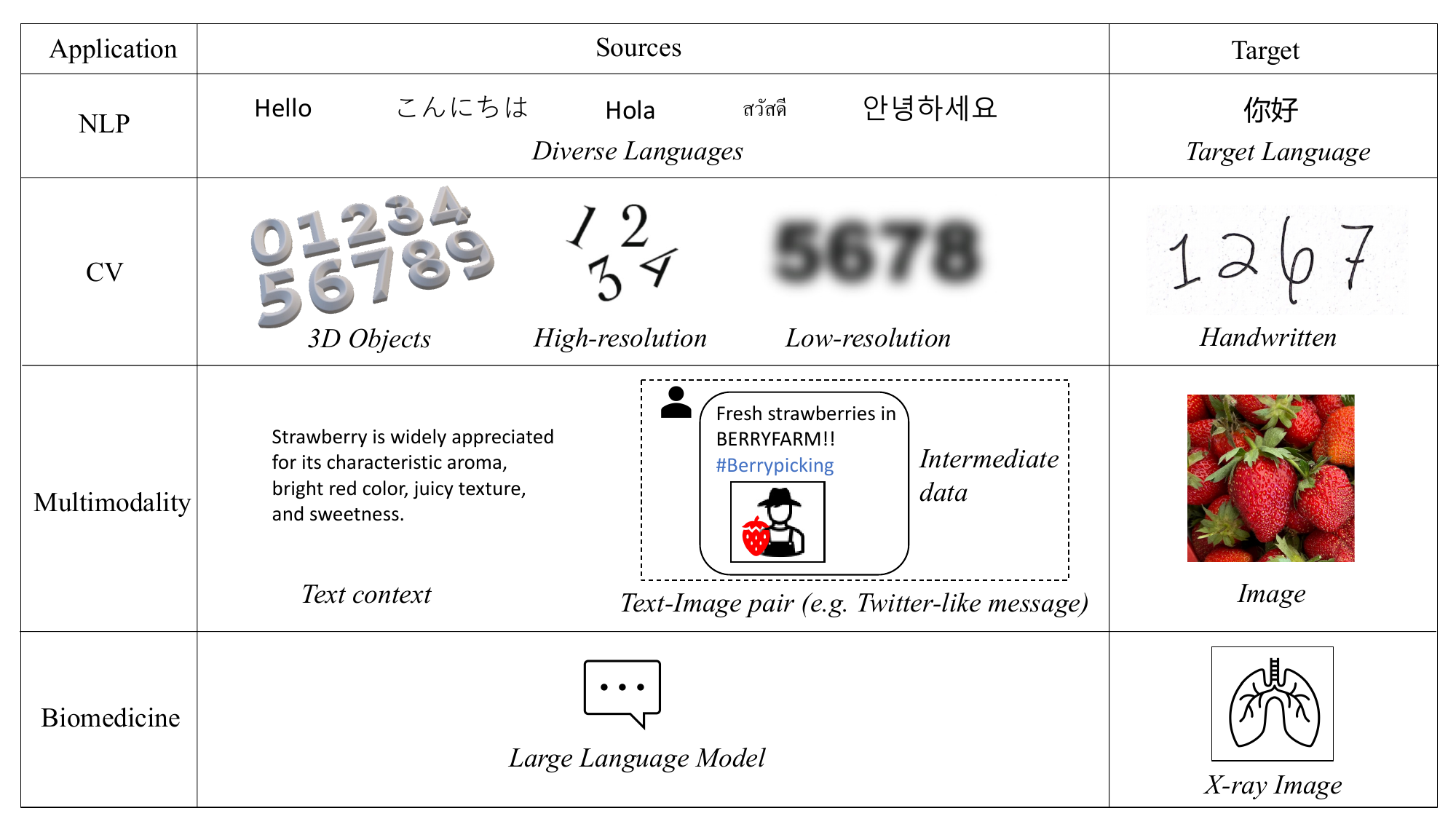}}
\caption{Heterogeneity in application scenarios}
\label{fig6}
\end{figure}

In this section, we will delve into the utilization of HTL methods in specific areas, including NLP, CV, Multimodality, and Biomedicine, as outlined in Table \ref{table:application} and illustrated in Figure  \ref{fig6}. Through a detailed examination of methods in each of these domains, we aim to uncover the challenges and progress across diverse application contexts. Additionally, we highlight prominent datasets for HTL research, providing comprehensive details and referencing the specific methods that employed them, as detailed in Table \ref{table:dataset}.

\subsection{Natural Language Processing}

Transfer learning has emerged as a valuable approach in NLP to address the challenge of scarce labeled data in specific scenarios \cite{ruder2019transfer}. In the context of object classification tasks, several methods \cite{SHDA-RF, MGLE, OTLMS} leverage information from various domains and apply it to target domains for classifying documents in 20 Newsgroups text collection dataset.

For sentiment analysis tasks, Multi-Domain Sentiment Dataset \cite{MDSD} contains Amazon product reviews for four different product categories: books, DVDs, electronics, and kitchen appliances. By selecting one of these domains as the target domain, HTL methods \cite{MGLE, SHFA, KPDA, HTLA, SHFR} can effectively transfer insights and expertise from the remaining categories, enhancing model robustness and accuracy in domain-specific sentiment analysis.

Obtaining labeled data can be particularly challenging in low-resource languages. Transfer learning has emerged as a valuable strategy to mitigate this challenge by facilitating knowledge transfer from well-resourced languages, such as English, to low-resource languages. For example, various methods \cite{SGW, WOS:000429325500003, WOS:000631201900035, WOS:000707695400001, HDAPA, STN, KPDA, HTLA, NMF, OHKT, CDLS, OHTHE, SHFR, CDSPP, DACoM, CWAN, MGLE, OTLMS} have been developed to enable this information transfer across languages. These methods utilize multilingual datasets like the Multilingual Reuters Collection Dataset \cite{amini2009learning} and the Multilingual Amazon Reviews Corpus \cite{MARC}, covering languages including English, French, German, Italian, Spanish, Japanese, and Chinese. By employing these datasets, models are able to capture universal contextual dependencies and linguistic patterns that are shared across languages, thereby enhancing performance in NLP tasks across diverse linguistic settings.

\subsection{Computer Vision}
Transfer learning is widely applied in CV for several reasons. Firstly, it facilitates the transfer of knowledge from pre-trained models on large-scale datasets, such as ImageNet, to new tasks or domains with limited labeled data. This process not only saves time but also conserves computational resources. Secondly, transfer learning leverages shared visual features among different CV tasks, enabling faster model development and improved performance. Lastly, it addresses the challenge of domain shift by adapting models to variations in lighting, viewpoint, or image quality, thereby enhancing their robustness and generalization across different visual environments. Overall, transfer learning accelerates training, improves performance, and enhances the applicability of CV in various domains, including image classification, object recognition, image segmentation, person re-identification,.

One of the widely recognized tasks in HTL within the field of CV is cross-domain object recognition. For this purpose, the commonly employed dataset is an amalgamation of the Office and Caltech-256 datasets. The Office dataset \cite{ITML} includes images sourced from three distinct origins: images obtained from Amazon, high-resolution images captured with a digital SLR camera, and lower-resolution images taken using a web camera \cite{WOS:000821313900005, WOS:000356863900003, WOS:000429325500003, WOS:000707695400001}. By integrating images from the Caltech-256 dataset, which forms the fourth category, the resultant Office + Caltech-256 dataset is compiled by selecting categories that overlap between both datasets \cite{WOS:000375360900012, jin2022heterogeneous, LPJT, CDLS, HDAPA, STN, DCA, NMF, DeepMCA, ICDM, CDSPP}.

In the broader field of CV, diverse datasets are utilized for specialized tasks. For example, the CIFAR-10 and CIFAR-100 datasets are essential in image classification tasks and are invaluable for assessing knowledge transfer across varied categories \cite{WOS:000564647800001}. The UCI dataset \cite{jain2000statistical}, particularly noted for tasks centered around handwritten digit recognition \cite{SCP}, has proven to be a reliable resource. Furthermore, a notable study \cite{WOS:000576289700036} examines the selection of 3D objects from renowned datasets such as NTU \cite{chen2003visual} and ModelNet40 \cite{wu20153d}, exploring knowledge transfer in this context. In the area of heterogeneous face recognition, datasets such as CASIA \cite{li2013casia}, NIVL \cite{bernhard2015near}, and the CMU Multi-Pie dataset \cite{gross2010multi} are frequently employed \cite{WOS:000462364200010, WOS:000595969000001, WOS:000356863900003}. These datasets collectively contribute to the exploration of knowledge transfer and transfer learning in CV applications.

\subsection{Multimodality}
When learning with multimodal data, aligning feature spaces effectively presents significant challenges. In these scenarios, HTL becomes invaluable. Its strength lies in its ability to harness auxiliary data as intermediaries, facilitating a smooth information flow between modalities and effectively bridging the gap between source and target domains.

Multimodal tasks often involve both images and text. For instance, consider the context of image classification as the target learning task, where a collection of text documents serves as auxiliary source data. In the research conducted in \cite{OHKT, wu2013cotransfer}, co-occurrence data, such as text-image pairs, serve as this intermediate data to establish a connection between the source and target domains. This type of data is often readily available and easily collected from social networks, providing a cost-effective solution for knowledge transfer. The representations of images can be enriched by incorporating high-level features and semantic concepts extracted from auxiliary images and text data \cite{HTLIC}.

Additionally, the NUS-WIDE dataset \cite{NUSWIDE} finds common applications in text-to-image classification tasks. This extensive dataset comprises 45 tasks, each composed of 1200 text documents, 600 images, and 1600 co-occurred text-image pairs \cite{WOS:000361685100003}. This dataset can be extended by incorporating images from the ImageNet dataset as in \cite{TNT} or text-image pairs extracted from ``Wikipedia Feature Articles'' \cite{Wikipedia} as demonstrated in studies like \cite{jin2022heterogeneous, WOS:000707695400001, STN, HDAPA, DeepMCA, OHKT, ICDM, CDSPP, SCT, CWAN}.

\subsection{Biomedicine}
Heterogeneity commonly exists in biomedicine: (a) Medical terminology undergoes continuous evolution, leading to the retirement of outdated terms and the introduction of novel ones. On occasions, these changes can be substantial, as exemplified by the transition from ICD-9 to ICD-10 coding systems; (b) The extensive adoption of electronic health record systems (EHRs) opens up substantial opportunities for deriving insights from routinely accumulated EHR data. However, the existence of distinct EHR structural templates and the utilization of local abbreviations for laboratory tests across various healthcare systems result in considerable heterogeneity among the collected data elements; (c) The potential of leveraging large language models and visual models in biomedicine may encounter challenges in effectively integrating and adapting to new data components, including medical terms, biomedical concepts (such as protein structures), and medical images.

Addressing this heterogeneity is crucial, and HTL strategies have evolved over time. Previous HTL approaches include basic data augmentation, incorporation of prior knowledge into the source Bayesian network \cite{ye2019transfer}, and a matrix projection method that only requires each source domain to share the empirical covariance matrix of the features \cite{wang2022survmaximin2}. Recent explorations have begun to augment large fundamental models with biomedical data types, such as protein 3D structures \cite{guo2023proteinchat}, drug compound graphs \cite{liang2023drugchat}, chest X-ray images \cite{liang2023xraychat}. These data types are typically processed using encoding and projection layers to convert them into compatible formats for large foundational models. The training procedures often employ a partial fine-tuning strategy.

\section{Discussion and Future Directions}

HTL has emerged as a transformative approach in the realm of machine learning, addressing the complexities associated with divergent feature spaces, data distributions, and label spaces between source and target domains. This work aimed to offer a comprehensive examination of HTL in light of the recent advancements, particularly those made post-2017. As evidenced by the survey, HTL methodologies have shown significant promise, especially in fields such as NLP, CV, Multimodality, and Biomedicine. It offers a robust mechanism to tackle the challenges faced in data-intensive fields across domains. The surveyed methods and techniques underscore their adaptability and versatility across a range of scenarios. After a thorough review of the existing techniques in HTL, we would like to highlight some key insights, opportunities, and challenges in the domain of HTL.

\paragraph*{Scarce labeled target data challenges:}
Real-world applications that need transfer learning often involve abundant labeled data in the source domain and limited data from the target domain. When the target domain lacks any labels, it poses significant challenges. Handling strategies include: (a) utilizing corresponding source-target pairs to match unlabeled target samples with samples in the source domain \cite{HTLA, DHTL}; (b) matching the marginal distributions of source and target features using MMD \cite{LPJT, HDAPA}; and (c) employing domain adversarial learning to reduce distribution discrepancies \cite{SymGAN}. By leveraging the readily available unlabeled data in the target domain, transfer learning can be facilitated more effectively.



\paragraph*{Method suitability varies by scenario:}
The suitability of HTL methods is greatly influenced by the specific application scenarios encountered. For instance, when source and target domains significantly differ and additional information is accessible (e.g., co-occurrence source-target sample pairs or intermediate data like tags for images and text), instance-based strategies prove highly effective. These methods are both intuitive and straightforward to implement. Conversely, when only source and target domain data are available, feature representation-based methods are advisable. Their flexibility and broad applicability make them ideal for diverse applications. Among feature-representation-based methods, feature augmentation techniques preserve domain-specific patterns, which can be advantageous when these patterns are critical to the task at hand. In situations when the availability of source data is constrained, model-based techniques offer a practical alternative. These methods enable knowledge transfer through pre-existing source models, ensuring data privacy and boosting computational efficiency by transferring only the model architecture or parameters instead of the entire dataset. 
Finally, real-world transfer scenarios often involve more complex situations, such as multi-source transfer \cite{ji2023prediction},  and online learning in the target domain \cite{sun2024online}. Consequently, there is a need for the development of more HTL methods tailored to these challenging scenarios.


\paragraph*{Advanced training methods develop domain-agnostic pre-trained models:}
In recent years, there has been a marked shift towards the use of pre-trained model-based methodologies. Among these, Large Language Models like the Generative Pre-trained Transformer stand out due to their remarkable capabilities. These models' architectures possess an extensive number of parameters, refined through comprehensive self-supervised multitask learning on vast text corpora. This reduces their reliance on domain-specific labeled data, thereby enhancing their adaptability across diverse downstream tasks. Further refinement through fine-tuning specialized datasets ensures that these models excel in targeted applications, paving the way for the development of more robust and sophisticated language comprehension systems. This could significantly influence future research in HTL.

\paragraph*{Domain disparities in multi-modality challenges:}
When the source and target domains differ not only in data distribution but also in modalities (e.g., transferring knowledge from text to image data or vice-versa), the challenges multiply \cite{zhen2020deep}. HTL in multimodal settings grapples with a series of obstacles: significant differences in feature spaces and data representations across modalities, a lack of shared feature space, and the risk of negative transfer when misleading or irrelevant source domain knowledge is applied to the target domain. Additionally, there are challenges in developing algorithms capable of effectively aligning and mapping representations from one modality to another, while retaining the salient and discriminative features crucial for the target task. Bridging the semantic gap between different modalities often requires innovative fusion techniques and domain adaptation strategies, 
necessitating a deeper understanding and the development of novel methodologies to ensure effective and meaningful knowledge transfer.

\paragraph*{Knowledge distillation  falters:}
In transfer learning, knowledge distillation is a pivotal technique \cite{hinton2015distilling, gou2021knowledge}.  
The process typically involves transferring insights from a complex "teacher" model to a simpler "student" model, assuming both operate within the same feature and label spaces.  However, its effectiveness diminishes in heterogeneous scenarios where feature and label spaces vary significantly between domains. This limitation stems from knowledge distillation's reliance on congruent data structures and tasks between the teacher and student models. When these tasks differ, the sophisticated abstractions learned by the teacher may not be relevant or could even negatively impact the student model's performance in its specific context. Thus, while effective for model simplification within homogeneous domains, knowledge distillation has not been extensively explored for HTL.


\paragraph*{Interpretability is vital:}
As the complexity of HTL grows, its ability to connect diverse domains also exposes complex interactions. These interactions, often deeply embedded within the transferred layers, can be obscure and non-intuitive. Given these complexities, maintaining interpretability is crucial for several reasons \cite{molnar2020interpretable, ahmad2018interpretable}. Firstly, it enhances the model's robustness by clarifying how transferred knowledge affects the learning process in the target domain. This understanding allows practitioners to fine-tune or adapt models more effectively. Secondly, interpretability is essential for diagnosing errors—whether they arise from biases or inaccuracies in the source domain or from faulty mappings during the transfer process. Lastly, from an ethical perspective, ensuring that the decision-making process is transparent and justifiable is critical, especially in sectors like healthcare, finance, and the judiciary. Without interpretability, the opaque nature of many complex HTL methods could lead to unintended consequences, undermining trust and potentially perpetuating biases.\\


As the machine learning landscape evolves, so too will the paradigms and techniques within HTL, requiring ongoing exploration, adaptation, and understanding.
First, addressing the challenges posed by unsupervised transfer learning scenarios when labels of the target domain is rare could unlock significant advancements, bridging the gap between abundant labeled source data and scanty labeled target data. Additionally, the advancement of multi-modal knowledge transfer techniques will be instrumental in navigating the complexities of disparate data domains and representations. Moreover, the burgeoning realm of pre-trained model methodologies, especially in the context of Large Language Models, offers significant opportunities for fine-tuning and adaptation across diverse tasks, underscoring the need for scalable, efficient, and more robust fine-tuning paradigms. Furthermore, knowledge distillation,  
though limited in the heterogeneous setting, may find resurgence through novel techniques that facilitate the transfer of knowledge across domains without the pitfalls of negative transfer. Lastly and may be the most important, the quest for model interpretability in HTL remains paramount. Future research should prioritize the development of frameworks that not only improve the transparency of these models but also enhance the effectiveness and ethical application. Such advancements will not only shape the trajectory of HTL but also bolster its real-world impact across various interdisciplinary domains towards the right direction.

\section{Conclusion}

Heterogeneous transfer learning (HTL) has become an essential tool in the modern landscape of machine learning, addressing the persistent challenge of data scarcity in real-world scenarios where source and target domains differ in feature or label spaces. 
This survey offers a comprehensive examination over 60 methods, categorizing them into data-based and model-based approaches. 
By systematically reviewing a wide range of recent methods, including instance-based, feature representation-based, parameter regularization, and parameter tuning techniques, we highlight the diversity of methodologies and their applications across various domains. Our comprehensive analysis of the underlying assumptions, calculations, and algorithms, along with a discussion of current limitations, offers valuable guidance for future research. This ensures that emerging HTL methods can address the identified gaps and advance the field. Moreover, by incorporating recent advancements like transformer-based models and multi-modal learning, we ensure that our survey reflects the latest developments and trends. This work not only bridges significant gaps in the literature but also serves as a crucial resource for researchers aiming to develop more robust and effective HTL techniques. The extensive coverage and critical insights offered by this survey are poised to stimulate further research and innovation in HTL, paving the way for its broader application and more significant impact in various real-world scenarios.




\section*{Acknowledgement}
Yiming Sun, Yuhe Gao, and Ye Ye were supported by the research grant R00LM013383 from the National Library of Medicine, National Institutes of Health. 

\printcredits

\section*{Declaration of Generative AI and AI-assisted Technologies in the Writing Process}
During the preparation of this work, the authors used ChatGPT in order to improve readability and language. After using this service, the authors reviewed and edited the content as needed and took full responsibility for the content of the publication.











\bibliographystyle{elsarticle-num}

\bibliography{ref}



\end{document}